\documentclass[11pt]{article}

\usepackage[final]{acl}

\usepackage{times}
\usepackage{latexsym}
\usepackage[T1]{fontenc}
\usepackage[utf8]{inputenc}
\usepackage{microtype}
\usepackage{inconsolata}

\usepackage{graphicx}
\usepackage[nolist]{acronym}
\usepackage{tabularx}
\usepackage{makecell}
\usepackage{multirow}
\usepackage{booktabs}
\usepackage{longtable}
\usepackage{caption}
\usepackage[most]{tcolorbox}

\newtcolorbox{takeawaybox}{
  enhanced,
  breakable,
  colback=gray!4,
  colframe=gray!55,
  boxrule=0.2pt,
  arc=2mm,
  left=0.2pt,
  right=0.2pt,
  top=0pt,
  bottom=0pt,
  before skip=4pt,
  after skip=4pt,
}

\title{Attribution, Citation, and Quotation: A Survey of Evidence-based Text Generation with Large Language Models}

\author{
  Tobias Schreieder$^1$ \and Tim Schopf$^{1,2}$ \and Michael Färber$^1$
  \\
  $^1$TU Dresden \& ScaDS.AI Dresden/Leipzig, Dresden, Germany \\
  $^2$National Institute of Informatics, Tokyo, Japan \\
  \texttt{\{tobias.schreieder, tim.schopf, michael.faerber\}@tu-dresden.de}
}

\begin{acronym}
    \acro{nli}[NLI]{natural language inference}
    \acro{nlp}[NLP]{natural language processing}
    \acro{ai}[AI]{artificial intelligence}
    \acro{qa}[QA]{Question answering}
    \acro{llm}[LLM]{large language model}
    \acro{rag}[RAG]{retrieval-augmented generation}
\end{acronym}

\begin{document}
\maketitle
\begin{abstract}
The increasing adoption of \acp{llm} has raised serious concerns about their reliability and trustworthiness. As a result, a growing body of research focuses on \textit{evidence-based text generation with \acp{llm}}, aiming to link model outputs to supporting evidence to ensure traceability and verifiability. However, the field is fragmented due to inconsistent terminology, isolated evaluation practices, and a lack of unified benchmarks. To bridge this gap, we systematically analyze 134 papers, introduce a unified taxonomy of evidence-based text generation with \acp{llm}, and investigate 300 evaluation metrics across seven key dimensions. Thereby, we focus on approaches that use citations, attribution, or quotations for evidence-based text generation. Building on this, we examine the distinctive characteristics and representative methods in the field. Finally, we highlight open challenges and outline promising directions for future work.
\end{abstract}

\section{Introduction} 
\label{sec:introduction}

Recent \acp{llm} have demonstrated remarkable capabilities in language understanding and generation \cite{NEURIPS2020_1457c0d6,NEURIPS2022_b1efde53}. Despite these advances, \acp{llm} continue to face challenges such as the tendency to generate hallucinations \cite{10.1145/3571730,10.1145/3703155} and their knowledge being limited to training data \cite{zhang-etal-2023-large,li2025memorizationvsreasoningupdating}.

\begin{figure}[ht!]
    \centering
    \includegraphics[width=\linewidth]{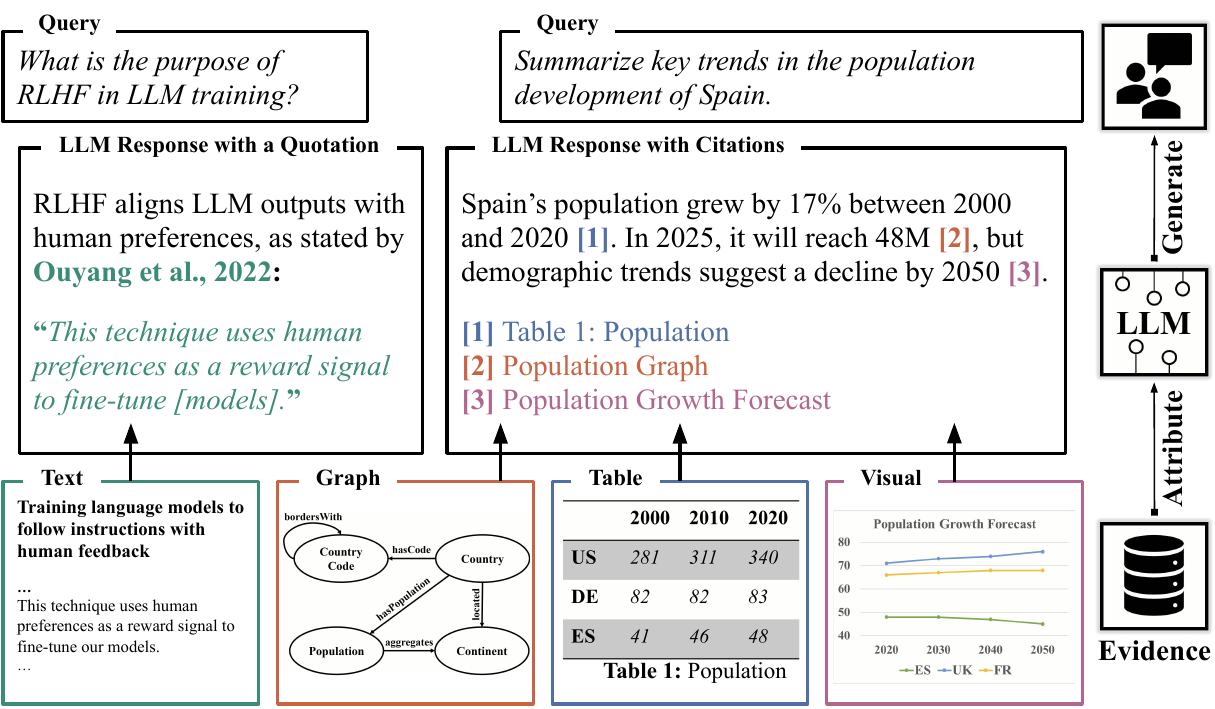}
    \caption{Illustration of evidence-based text generation with \acp{llm} with various citation modalities and styles.}
    \label{fig:overview}
\end{figure}

To address these limitations and increase trust, an emerging line of research focuses on generating text that is traceable to supporting evidence, allowing verification of \ac{llm}-generated content~\cite{huang-chang-2024-citation}. However, despite growing interest, there is no shared understanding of the field, as prior works have used varied terminology to describe similar research efforts. For instance, while \ac{rag} has gained prominence in recent years, our survey identifies it as just one of seven closely related approaches. Consequently, research efforts in this area are highly fragmented, with studies typically aiming to cite (used in 75\% of papers), attribute (62\%), or quote (13\%) evidence.\footnote[1]{Some studies use multiple terminologies and approaches.} Generating text with citations involves the insertion of citation markers that explicitly reference supporting evidence sources \cite{gao-etal-2023-enabling}. Attributed text generation constitutes a broader notion that generally focuses on linking generated content back to the underlying sources used for grounding \cite{slobodkin-etal-2024-attribute}. Further, generating text with quotes focuses on incorporating excerpts from evidence sources into the generated text \cite{menick2022teachinglanguagemodelssupport}. While these research efforts differ in focus, they share the common paradigm of ``\textit{evidence-based text generation with \acp{llm}}'', where \acp{llm} generate texts, accompanied by explicit references that make the outputs traceable to supporting evidence. Figure~\ref{fig:overview} illustrates this task across different scenarios.

Given the rapid advancement of \acp{llm} and diverse research on evidence-based text generation, a wide range of approaches has emerged, calling for consolidation. However, no existing study offers a comprehensive and systematic review of the full research landscape in this area. To address this gap, we conducted an extensive survey, categorizing key concepts, identifying trends, and outlining promising directions for future work. To the best of our knowledge, this is the first survey dedicated to this paradigm. Our analysis of 134 papers, 300 evaluation metrics, 19 frameworks, 231 datasets, and 11 benchmarks serves as a valuable resource for understanding and navigating the field. We make our annotated dataset available in a public repository.\footnote[2]{Dataset: \url{https://github.com/faerber-lab/AttributeCiteQuote}}

The main contributions of this study are:
\begin{enumerate}
    \item We provide the first taxonomy of evidence-based text generation with \acp{llm}.
    \item We review 300 evaluation metrics, classify them by seven dimensions and six methods, and identify common benchmarks.
    \item We outline emerging research trends, key limitations, and promising future directions.
\end{enumerate}

\section{Related Work}
\label{sec:related_work}

Most existing surveys address related but distinct topics, such as \ac{llm} hallucinations \cite{10.1145/3571730,zhang2023sirenssongaiocean,sahoo-etal-2024-comprehensive,10.1145/3703155}, knowledge-enhanced text generation \cite{10.1145/3512467}, grounding capabilities of \acp{llm} \cite{lee-etal-2024-well,qiu-etal-2024-large,jokinen-2024-need}, generative information retrieval \cite{Li2025SurveyGenerativeIR}, and \ac{rag} \cite{10.1145/3637528.3671470,ARSLAN20243781,gao2024retrievalaugmentedgenerationlargelanguage,Chen_Lin_Han_Sun_2024}. While these works study important components of evidence-based text generation, they focus on isolated aspects such as factuality, retrieval, or grounding, without providing a unified perspective.

Research on AI-generated plagiarism detection is related, as both address attribution for semantic reuse~\cite{Pudasaini2025AIPlagiarism,wu-etal-2025-survey}. However, plagiarism detection focuses on identifying uncredited content, often from a legal or ethical perspective, emphasizing post-hoc analysis. In contrast, evidence-based text generation with \acp{llm} aims to generate text supported by verifiable sources through explicit citation.

Prior literature on evidence-based text generation with \acp{llm} remains limited in scope. In their position paper, \citet{huang-chang-2024-citation} emphasize the importance of citation mechanisms but do not systematically review prior work. \citet{li2023surveylargelanguagemodels} discuss early research on \ac{llm} attribution but do not cover the broader paradigm of evidence-based text generation, lack a systematic literature search, and are already outdated, as over 75\% of studies in our dataset were published after 2023. In contrast, our survey systematically reviews the full landscape of evidence-based text generation with \acp{llm}, covering attribution approaches, citation characteristics, tasks, and evaluation resources.

\section{Evidence-based Text Generation}
\label{sec:evidence_based_text_generation}

\begin{figure*}[ht!]
    \centering
    \includegraphics[width=\linewidth]{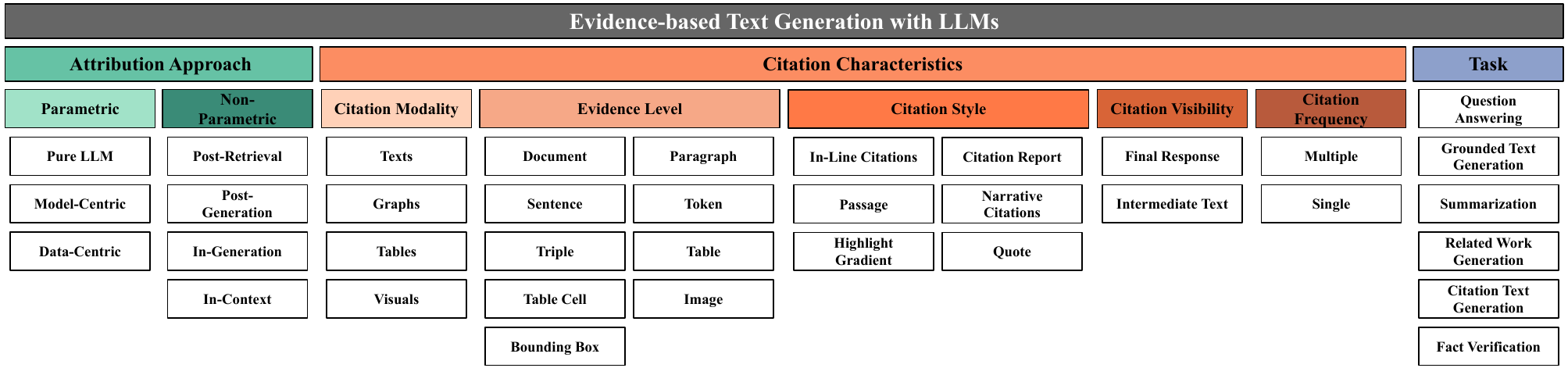}
    \caption{Multidimensional taxonomy of evidence-based text generation with \acp{llm}. The taxonomy categorizes papers along three independent dimensions: attribution approach, citation characteristics, and task, which together capture the core design choices of evidence-based text generation. Table~\ref{tab:classification_papers} in Appendix lists all annotated papers.}
    \label{fig:taxonomy}
\end{figure*}

This section presents the findings of our literature review. We conducted a systematic mapping study following the PRISMA protocol, yielding 805 deduplicated papers, of which 134 were identified as relevant through manual screening. Figure~\ref{fig:taxonomy} presents the multidimensional taxonomy we developed to characterize evidence-based text generation with \acp{llm}, using a facetted classification approach~\cite{Crowston2004FacettedClassification}. For each taxonomy dimension, we provide a descriptive overview along with key findings and future directions, while methodological details and extended analyses are reported in Appendices~\ref{sec:methodology}, \ref{sec:research_landscape}, and~\ref{sec:additional_results}.

\begin{figure}[h]
    \centering
    \includegraphics[width=\linewidth]{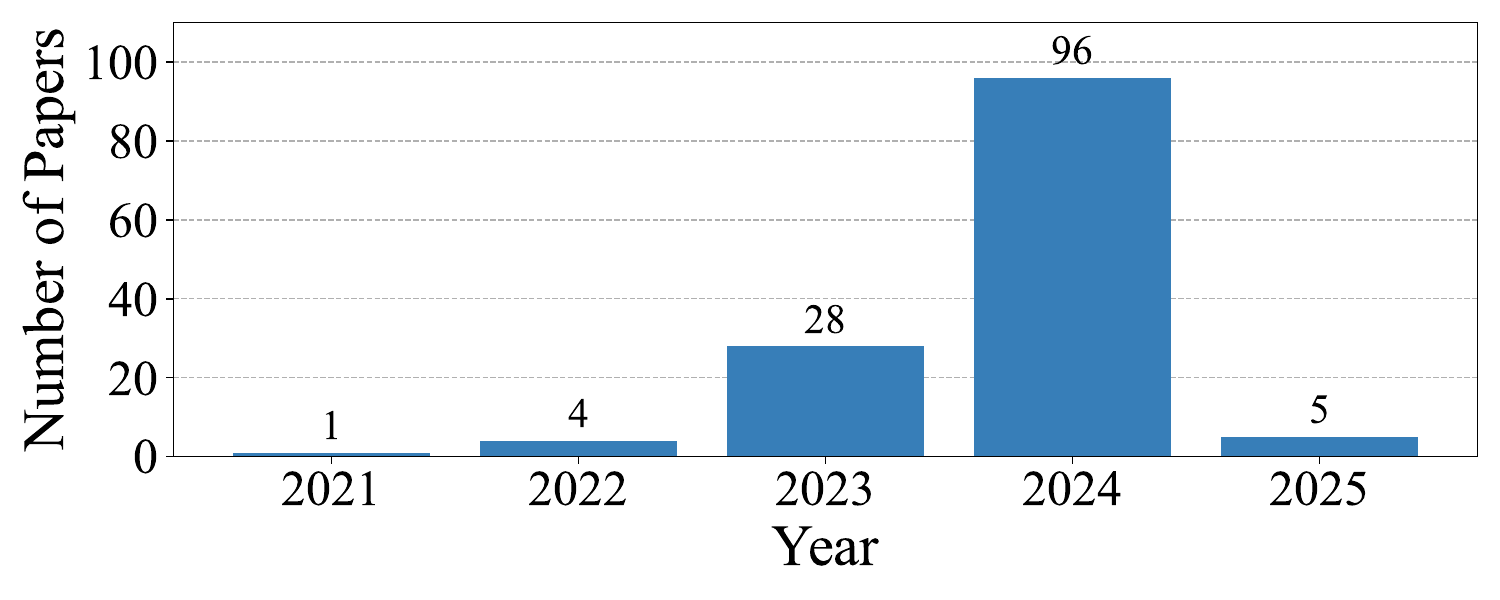}
    \caption{Number of studies per year.}
    \label{fig:paper_year}
\end{figure}

We analyze the number of published papers per year as a proxy for research interest. Figure~\ref{fig:paper_year} shows that after a few studies in 2021–2022, papers increased to 28 in 2023 and surged to 96 in 2024, a 3.4-fold rise. Over 75\% of studies were published after 2023, highlighting limitations of earlier surveys such as \citet{li2023surveylargelanguagemodels}, which omit much of the recent literature. As our search was conducted in February 2025, the dataset includes only five papers from that year. Given the trend, we anticipate continued growth, underscoring sustained interest in evidence-based text generation with \acp{llm}.

\subsection{Attribution Approach}
\label{sec:attribution_approach}

\citet{huang-chang-2024-citation} categorize attribution into parametric and non-parametric approaches, distinguished by the nature of knowledge used during text generation. \textit{Parametric approaches} rely on knowledge encoded within model parameters during training and are suited for analyzing knowledge within \acp{llm} by enabling attribution without reliance on external sources, supporting explainability. In contrast, \textit{non-parametric approaches} incorporate external sources at inference time and are the predominant choice in current evidence-based text generation systems, as they enable the use of explicit and up-to-date evidence. We extend this categorization to include approaches outside the original framework. In total, we identify 126 non-parametric and 25 parametric attribution approaches, with post-retrieval being most common. A detailed distribution across tasks is shown in Figure~\ref{fig:attribution_per_task} in Appendix.

\subsubsection{Parametric Attribution}
\label{sec:parametric_attribution}

Previous work treats parametric attribution as a single category~\cite{huang-chang-2024-citation, li2023surveylargelanguagemodels}, while our review identifies three types.
\textbf{Pure \acp{llm}} rely on the inherent attribution capabilities of existing \acp{llm} without altering their architecture, training procedures, or underlying data~\cite{byun-etal-2024-reference, Zuccon2023ChatGPTCitation}. \textbf{Model-Centric} approaches aim to improve attribution by modifying the \ac{llm} architecture or training objectives \cite{chu-etal-2025-towards, Khalifa2024SourceAwareTraining}. Finally, \textbf{Data-Centric} attribution curates, augments, or synthetically generates data without changing \ac{llm} architectures or internal behavior \cite{li-etal-2024-improving-attributed, huang-etal-2024-learning}.

\begin{takeawaybox}
\textbf{Takeaways.} Parametric attribution remains largely underexplored, with only 25 papers in our study, most of which (72\%) focus on evaluating attribution behavior of pure \acp{llm}. Model- and data-centric approaches receive limited attention, with few methods targeting even widely studied tasks and no clear growth trend (see Figure~\ref{fig:attribution_per_task}), reflecting the technical difficulty and limited scalability of current approaches. While evaluation of pure \acp{llm} remains important, future work could increasingly focus on model- and data-centric approaches to strengthen the intrinsic ability of \acp{llm} to generate attributable text. This is essential for understanding model-internal knowledge and data provenance, enabling more effective analysis of hallucinations, and supporting assessments of privacy and copyright.
\end{takeawaybox}

\subsubsection{Non-Parametric Attribution}
\label{sec:non_parametric_attribution}

Non-parametric attribution relies on the integration of external evidence. \citet{li2023surveylargelanguagemodels} distinguish between \textit{post-retrieval} and \textit{post-generation} approaches. We extend this categorization with the two classes \textit{in-generation} and \textit{in-context}.

\textbf{Post-Retrieval} attribution, also known as \textit{pre-hoc} \cite{huang-chang-2024-citation}, refers to a class of methods which retrieve relevant external information before text generation \cite{li2023surveylargelanguagemodels}. The retrieved content is incorporated as additional context to condition the \ac{llm}'s response. A prominent architecture in this category is \ac{rag}, wherein the \ac{llm} is instructed to base its output solely on the retrieved documents. However, standard \ac{rag} does not inherently support attribution, and must be extended with mechanisms for providing explicit references to the retrieved source documents.

\textbf{Post-Generation} attribution, also known as \textit{post-hoc} \cite{huang-chang-2024-citation}, first generates a response and then retrieves relevant evidence based on the \ac{llm} output \cite{li2023surveylargelanguagemodels}. These approaches resemble classical citation recommendation~\cite{faerber-cite-survey} by retrieving sources for each generated claim, leveraging evidence retrieval on both the user input (e.g., a question) and the \ac{llm} output to attribute each generated claim.

\textbf{In-Generation} attribution is a recent paradigm in which \acp{llm} dynamically determine the need for additional evidence and retrieve it during generation~\cite{asai2024selfrag, li2024nearest}. This contrasts with traditional post-retrieval or post-generation methods by tightly integrating retrieval decisions into the generation process.

\textbf{In-Context} attribution differs from previous approaches, as it does not require retrieval. Instead, the user explicitly provides evidence as part of the prompt, embedding it directly into the context~\cite{cohen-wang2024contextcite,zhang-etal-2025-longcite}.

\begin{takeawaybox}
\textbf{Takeaways.} Non-parametric attribution dominates the literature, comprising 126 works in our study. Post-retrieval attribution is the most prevalent paradigm, covering 58\% of non-parametric approaches. This prevalence reflects the practical effectiveness of \ac{rag} pipelines. Post-generation attribution remains comparatively limited (18\%) and is concentrated primarily in question answering (see Figure~\ref{fig:attribution_per_task} in Appendix). In-context attribution (20\%) is task-dependent, mainly appearing in settings where evidence is provided as input, such as summarization and citation text generation. In contrast, in-generation attribution remains underexamined (4\%) but represents a promising direction for reducing the limitations of rigid evidence retrieval for non-parametric attribution.
\end{takeawaybox}

\subsection{Citation Characteristics}
\label{sec:citation_characteristics}

The five citation characteristics focus on the nature and presentation of evidence.

\subsubsection{Citation Modality}
\label{sec:citation_modality}

Although evidence-based text generation requires \acp{llm} to generate text, the modality of the underlying evidence can differ. We identify four citation modalities, with the most prevalent being \textbf{texts}, which covers all forms of unstructured textual data~\cite{malaviya-etal-2024-expertqa}. Studies also cite \textbf{graphs} as structured representations of entities and relations~\cite{he-etal-2025-evaluating-improving}, \textbf{tables} as data organized in rows and columns~\cite{mathur-etal-2024-matsa}, and \textbf{visuals} such as images~\cite{ma-etal-2025-visa}.

\begin{takeawaybox}
\textbf{Takeaways.} Text overwhelmingly dominates citation modalities in current work (96\% of studies). Citing non-textual evidence remains largely unexplored, pointing to multimodality as an important direction for future research.
\end{takeawaybox}

\subsubsection{Evidence Level}
\label{sec:evidence_level}

Each citation modality can be cited at different levels of granularity, which we refer to as the evidence level. For \textit{text}, cited evidence can correspond to a full \textbf{document}, such as a scientific article~\cite{Anand2023CitationTextGeneration}, a \textbf{paragraph}, for example a retrieved chunk~\cite{gao-etal-2023-enabling}, a single \textbf{sentence}~\cite{xu-etal-2025-aliice}, or even individual \textbf{tokens}~\cite{phukan-etal-2024-peering}. For \textit{graphs}, evidence is cited at the level of a \textbf{triple}~\cite{li-etal-2024-towards-verifiable}, while no approach cites entire knowledge graphs. The \textit{tables} modality includes evidence at the level of a \textbf{table}~\cite{suri-etal-2025-visdom} or individual \textbf{table cells}~\cite{mathur-etal-2024-matsa}. Similarly, \textit{visual} evidence is cited either at the level of an \textbf{image}~\cite{suri-etal-2025-visdom} or a \textbf{bounding box}~\cite{ma-etal-2025-visa}.

\begin{takeawaybox}
\textbf{Takeaways.} The evidence level is dominated by coarse-grained texts, with document- and paragraph-level evidence accounting for the majority of studies (43\% and 40\%). As shown in Figure~\ref{fig:evidence_level_trend_analysis} in Appendix, these levels emerged earlier and continue to dominate the literature, while finer-grained evidence such as sentences and tokens remain less prevalent (12\% and 2\%), but exhibit stronger growth. Evidence levels for non-textual modalities remain underexplored, limiting conclusions about their granularity.
\end{takeawaybox}

\subsubsection{Citation Style}
\label{sec:citation_style}

As illustrated in Figure~\ref{fig:citation_styles} in Appendix, we identify six citation styles that differ in how evidence is presented to users. \textbf{In-line citations} place references directly after citation-worthy claims~\cite{huang-etal-2024-training}. \textbf{Citation reports} provide a separate list of references alongside the \ac{llm} output~\cite{bohnet2023attributedquestionansweringevaluation}. Some approaches display only the \textbf{passage} retrieved or used during generation, particularly in evaluation settings~\cite{muller-etal-2023-evaluating}. \textbf{Narrative citations} integrate references into the natural flow of the generated text to improve contextual clarity~\cite{shaier-etal-2024-adaptive}. \textbf{Highlight gradients} visually indicate source influence by coloring relevant tokens or sentences in the output and the supporting evidence~\cite{Do2024facilitatinghumanllmcollaborationfactuality}. Finally, \textbf{quotes} embed verbatim excerpts from the evidence directly into the generated response~\cite{xiao2025quillquotationgenerationenhancement}.

\begin{takeawaybox}
\textbf{Takeaways.} In-line citations dominate current practice, appearing in 62\% of studies (see Figure~\ref{fig:citation_style_numbers} in Appendix). For user-facing applications, citation style strongly affects verifiability: in-line citations enable claim-level verification of \ac{llm}-generated text, while styles such as highlight gradients or quotes additionally allow users to directly identify the supporting evidence spans.
\end{takeawaybox}

\subsubsection{Citation Visibility}
\label{sec:citation_visibility}

Citation visibility determines whether citations are shown to users. Most approaches include citations in the \textbf{final response}, enabling users to trace claims to their sources~\cite{gao-etal-2023-enabling}. In contrast, some approaches generate citations only in an \textbf{intermediate text}, where citations are used internally by the \ac{llm} rather than exposed for direct user traceability~\cite{fang-etal-2024-hgot}.

\begin{takeawaybox}
\textbf{Takeaways.} Citation visibility is predominantly user-facing, with 91\% of studies providing citations in the final response. Intermediate citation generation is rare and appears only in question answering and grounded text generation.
\end{takeawaybox}

\subsubsection{Citation Frequency}
\label{sec:citation_frequency}

Citation frequency captures the number of citations assigned to an \ac{llm}-generated claim. Existing approaches differ in whether they provide a \textbf{single} citation~\cite{shaier-etal-2024-adaptive} or \textbf{multiple} citations per claim~\cite{Khalifa2024SourceAwareTraining}.

\begin{takeawaybox}
\textbf{Takeaways.} Most studies support multiple citations per claim (64\%), particularly in non-parametric attribution settings where retrieval naturally enables citing several sources. In contrast, parametric approaches are often constrained by model architecture and may support only single citations. This suggests an open design space around when multiple citations are beneficial versus when single citation strategies suffice.
\end{takeawaybox}

\subsection{Task}
\label{sec:task}

We identify six frequent tasks in evidence-based text generation with \acp{llm}. \textbf{\ac{qa}}~\cite{gao-etal-2023-enabling} evaluates an \ac{llm}’s ability to answer questions by generating answers grounded in evidence. \textbf{Grounded text generation}~\cite{cheng-etal-2025-coral} captures more general generation settings, such as open-ended text or dialogue. \textbf{Summarization}~\cite{deng-etal-2024-webcites} assesses whether generated summaries attribute source documents, while \textbf{fact verification}~\cite{buchmann-etal-2024-attribute} focuses on determining the correctness of claims with respect to supporting evidence. Finally, \textbf{citation text generation}~\cite{Anand2023CitationTextGeneration} focuses on generating citation contexts in scientific writing where references are inserted into an existing manuscript. In contrast, \textbf{related work generation}~\cite{byun-etal-2024-reference} targets broader document-level synthesis, generating literature overviews that organize and relate prior works.

\begin{takeawaybox}
\textbf{Takeaways.} The task landscape of evidence-based text generation with \acp{llm} reflects a gradual expansion from early tasks such as question answering and grounded text generation toward more specialized settings, including citation text generation and fact verification. However, since most approaches and evaluation practices were developed around these dominant tasks, they may inadequately capture the requirements of newer tasks that involve reasoning over multiple sources or operate at different evidence levels. Future research could examine how existing approaches and evaluation practices transfer across tasks, and where task-specific adaptations are required.
\end{takeawaybox}

\section{LLM Integration}
\label{sec:llm_integration}

This section provides a complementary analysis of how \acp{llm} are operationalized in evidence-based text generation. In practice, \acp{llm} can be integrated at multiple stages of a system beyond text generation, for example to support attribution, citation generation, or task-specific requirements. Because these integration mechanisms span multiple taxonomy dimensions, we analyze \ac{llm} integration separately to preserve conceptual clarity. Across the reviewed literature, we identify two complementary integration strategies: \textit{training} and \textit{prompting}. Training-based approaches modify model behavior through pretraining or fine-tuning, whereas prompting-based approaches guide the model at inference time through structured inputs.

\subsection{Training}
\label{sec:training}

We annotate whether and how studies modify \acp{llm} through training, distinguishing between pretraining and fine-tuning. Few approaches employ \textbf{pretraining} either to incorporate attribution-specific objectives or to adapt models to broader settings. Several works treat pretraining as a necessary component for parametric attribution~\cite{Khalifa2024SourceAwareTraining, lu-etal-2025-wasa}, while others leverage it to support multilingual or multimodal scenarios~\cite{Abbas2025Fanar, patel-etal-2024-towards}.
More commonly, studies apply fine-tuning to adapt \acp{llm} to evidence-based text generation tasks. Fine-tuning is used to improve generation quality and attribution behavior, most often through \textbf{supervised fine-tuning}~\cite{li-etal-2024-improving-attributed, ye-etal-2024-effective}. In contrast, \textbf{self-supervised fine-tuning}~\cite{huang-etal-2024-advancing, chen2023purrefficientlyeditinglanguage} and \textbf{reinforcement learning}~\cite{huang-etal-2024-training, huang-etal-2024-learning} appear less frequently.

\begin{takeawaybox}
\textbf{Takeaways.} Training-based integration is used selectively in evidence-based text generation (45\% of studies). It is most common in model- and data-centric attribution, where attribution behavior is embedded in the \ac{llm} through fine-tuning or pretraining. Overall, fine-tuning primarily improves attribution quality but is also used for task adaptation when prompting alone is insufficient, indicating that training serves targeted methodological needs rather than a universal solution. This selective use highlights open questions about the limits of prompting, specifically which attribution behaviors can be achieved through prompting alone and which require parameter-level adaptation.
\end{takeawaybox}

\subsection{Prompting}
\label{sec:prompting}

In practice, the majority of studies rely on prompting to steer \acp{llm} toward evidence-based text generation without modifying \ac{llm} parameters. Common prompting techniques include \textbf{zero-shot}, \textbf{few-shot}, and \textbf{chain-of-thought} prompting, which are used to provide task instructions, in-context examples, or explicit reasoning steps~\cite{tahaei-etal-2024-efficient, Ateia2024BioRAGent, li-etal-2024-citation}. These techniques are frequently combined, for example by pairing few-shot demonstrations with explicit reasoning steps~\cite{shaier-etal-2024-adaptive}.
Beyond general-purpose prompting, several studies introduce strategies designed to improve citation behavior. These include \textbf{chain-of-citation} and \textbf{chain-of-quote} prompting~\cite{li-etal-2024-making}, which encourage explicit alignment between reasoning steps and cited evidence, as well as \textbf{conflict-aware} prompting~\cite{Patel2024FactualityOrFiction}. Additional prompting strategies are summarized in Table~\ref{tab:classification_papers} in Appendix.

\begin{takeawaybox}
\textbf{Takeaways.} Prompting is the predominant approach for operationalizing \acp{llm} in evidence-based text generation (78\% of studies). Most approaches rely on standard prompting strategies to guide the model behavior without parameter updates, reflecting the flexibility and low overhead of inference-time control. More specialized prompting strategies explicitly target attribution quality. Overall, prompting remains the state of the art for non-parametric attribution. The diversity of prompting strategies highlights the need for standardized templates in this domain.
\end{takeawaybox}

\section{Evaluation}
\label{sec:evaluation}

\begin{figure*}[ht!]
    \centering
    \includegraphics[width=\linewidth]{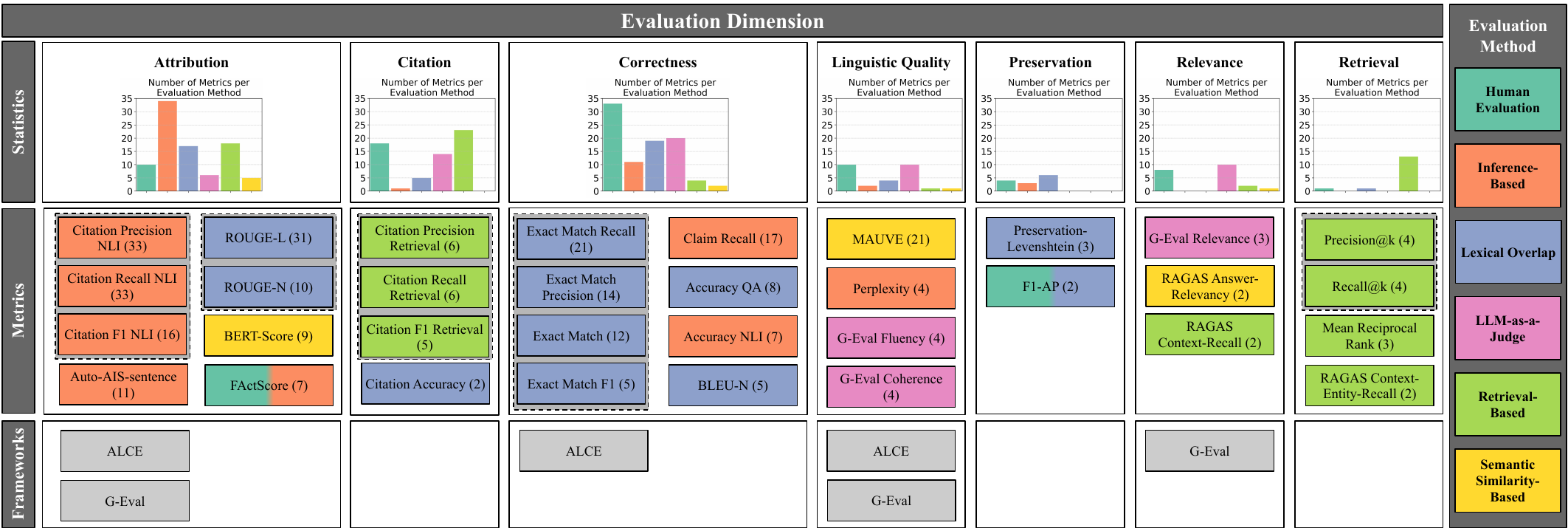}
    \caption{Frequently reused evaluation metrics and frameworks for evidence-based text generation. Numbers in parentheses indicate how many studies used each metric. Metrics grouped by dashed lines represent complementary metrics, recommended being used together. Additional, less reused metrics are listed in Appendix \ref{sec:additional_metrics}.}
    \label{fig:evaluation}
\end{figure*}

This section provides an overview of evaluation approaches for evidence-based text generation with \acp{llm}. In total, we identified 300 distinct metrics, each targeting different aspects of evaluation. Figure~\ref{fig:evaluation} offers a structured overview of frequently reused metrics categorized by \textit{evaluation method} and \textit{evaluation dimension}. We define \textit{reused} resources as those employed at least twice among surveyed studies. We observe that only two frameworks are reused across studies, namely ALCE \cite{gao-etal-2023-enabling} and G-Eval \cite{liu-etal-2023-g}. Overall, our survey comprises 19 frameworks (Table~\ref{tab:frameworks}), 11 benchmarks (Table~\ref{tab:benchmarks}), and 231 datasets (Table~\ref{tab:datasets}), detailed in Appendix~\ref{sec:evaluation_resources}.

\subsection{Evaluation Methods}
\label{sec:evaluation_methods}

The evaluation methods characterize the underlying strategies used to compute evaluation metric scores. Our initial categorization was derived from \citet{dziri-etal-2022-evaluating} and iteratively refined and extended during data annotation.
\textbf{Human evaluation} relies on human judges who rate \ac{llm}-generated texts along predefined criteria.
\textbf{Inference-based} metrics use \ac{nli} models to assess whether an \ac{llm}-generated text is entailed by a reference, while \textbf{lexical overlap} metrics focus on surface-level word matching.
\textbf{\ac{llm}-as-a-judge} metrics automatically assess the quality of texts generated by other \acp{llm}.
\textbf{Retrieval-based} metrics evaluate how effectively relevant evidence is incorporated by comparing retrieved evidence with ground truth.
\textbf{Semantic similarity-based} metrics assess the similarity between generated and reference texts using dense vector embeddings. More details are described in Table~\ref{tab:evaluation_method} in Appendix~\ref{sec:evaluation_resources}.

\subsection{Evaluation Dimensions}
\label{sec:evaluation_dimensions}

Evaluation dimensions specify which aspects of evidence-based text generation are evaluated.

\textbf{Attribution.} These metrics evaluate whether an \ac{llm}-generated output can be attributed to sources, without requiring citations or assessing factual correctness. Attribution is among the most critical and widely studied evaluation dimensions and is predominantly assessed via inference-based methods. The \textit{Citation \ac{nli}} metric cluster~\cite{gao-etal-2023-enabling} uses \ac{nli} models to determine whether \ac{llm} outputs are supported by sources and reports \textit{precision}, \textit{recall}, and $F_{1}$ scores. Related inference-based metrics include \textit{Auto-AIS-sentence}~\cite{gao-etal-2023-rarr}, which extends AIS~\cite{rashkin-etal-2023-measuring} to estimate the proportion of attributable sentences, and \textit{FActScore}~\cite{min-etal-2023-factscore}, which decomposes outputs into atomic facts and verifies their support via \ac{nli} or human annotation. Lexical overlap metrics such as the \textit{ROUGE} cluster~\cite{lin-2004-rouge} assess attribution by measuring surface-level overlap between generated text and evidence, while semantic similarity–based metrics, including \textit{BERTScore}~\cite{Zhang_2020BERTScore}, compare dense embeddings of \ac{llm} texts and references.

\textbf{Citation.} Metrics in this dimension assess whether an \ac{llm}-generated output cites appropriate evidence. Unlike attribution metrics, which focus on whether content can be traced to evidence, citation metrics emphasize the correctness and completeness of citations and are predominantly evaluated using retrieval-based methods. The most frequently adopted approach is the \textit{Citation Retrieval} metric cluster~\cite{deng-etal-2024-webcites}. These metrics compare cited sources in \acp{llm}-generated texts against a gold standard to assess whether \ac{llm}-generated citations support the content and are cited appropriately, reporting \textit{precision}, \textit{recall}, and $F_{1}$ scores. In addition, \textit{Citation Accuracy}~\cite{shaier-etal-2024-adaptive} evaluates whether citation strings in \ac{llm} outputs correspond to correct sources by matching them against a gold standard.

\textbf{Correctness.} This dimension evaluates whether an \ac{llm}-generated text is semantically accurate, such as generating answers without hallucinations or summaries without omitting key content. Correctness is predominantly assessed through human evaluation, reflecting limitations of automated metrics in capturing semantic accuracy. However, due to their heterogeneous implementation, human evaluation metrics are not frequently reported in Figure~\ref{fig:evaluation}. Among automated metrics, lexical overlap is widely used. The \textit{Exact Match} cluster, including \textit{recall}, \textit{precision}, $F_{1}$, and \textit{accuracy}, is used in factoid-style \ac{qa}~\cite{gao-etal-2023-enabling}, while \textit{BLEU-N} measures n-gram overlap with ground-truth texts~\cite{Papineni2022BLEU}. In multiple-choice \ac{qa} settings, correctness is evaluated using \textit{Accuracy QA}~\cite{chu-etal-2025-towards}. For long-form generation, inference-based metrics and \ac{llm}-as-a-judge approaches are increasingly employed, including \textit{Claim Recall}, which uses \ac{nli} models to verify whether generated outputs entail gold claims~\cite{gao-etal-2023-enabling}, and \textit{Accuracy NLI}, which assesses factual consistency between texts~\cite{malaviya-etal-2024-expertqa}.

\textbf{Linguistic Quality.} These metrics capture aspects such as fluency, referring to the naturalness and grammatical correctness of text, and coherence, which reflects the logical flow and consistency of ideas. Most metrics rely on human evaluation or the \ac{llm}-as-a-judge paradigm, including \textit{G-Eval Fluency} and \textit{G-Eval Coherence}~\cite{liu-etal-2023-g}. In addition, \textit{MAUVE}~\cite{pillutla2021mauve} is a widely used semantic similarity–based metric that assesses fluency and coherence by comparing distributions of generated and reference texts. The inference-based metric \textit{Perplexity}~\cite{liu-etal-2021-learning} measures a model’s uncertainty in next-token prediction, with lower values typically indicating more fluent text.

\begin{table*}[ht!]
\centering
\small
\renewcommand{\arraystretch}{1.25}
\setlength{\tabcolsep}{4pt}

\begin{tabular}{@{}%
  >{\arraybackslash}p{0.11\textwidth}%
  >{\arraybackslash}p{0.28\textwidth}%
  >{\arraybackslash}p{0.57\textwidth}%
@{}}
\toprule
\textbf{\makecell[l]{Evaluation\\Dimension}} & \textbf{When to Use} & \textbf{Explanation} \\
\midrule

\multicolumn{3}{@{}l@{}}{\textbf{Core}} \\
\addlinespace[0.3em]

Attribution
& When no annotated evidence is available
& Attribution is used when gold evidence is not provided (e.g., question answering with only ground-truth questions and answers). \ac{llm}-generated texts are compared against retrieved sources to determine whether claims can be supported by external evidence. \\

Citation
& When annotated ground-truth evidence is available
& Applicable when evidence is explicitly annotated (e.g., question–answer–evidence triples), enabling direct evaluation of whether the approach cites correct sources. In such settings, citation can be evaluated instead of, or in addition to, attribution. \\

Correctness
& Always
& Assesses factual accuracy of generated content. Correctness is fundamental to evidence-based text generation with \acp{llm} and should be evaluated regardless of task, system design, or evidence availability. \\

\midrule
\multicolumn{3}{@{}l@{}}{\textbf{Contextual}} \\
\addlinespace[0.3em]

Linguistic Quality
& When the \ac{llm} is modified, or when linguistic quality is critical for the use case
& Linguistic quality is relevant when the \ac{llm} generation process is altered (e.g., through pretraining or fine-tuning) or when high-quality language is essential for the application. \\

Preservation
& When a post-generation attribution approach is applied
& Preservation measures how much revised outputs deviate from the original generation. This dimension is particularly important for post-generation attribution approaches, which use external evidence to revise text generated by an \ac{llm}. \\

Relevance
& When the approach is deployed in user-centric settings
& Evaluates how well the \ac{llm}-generated output aligns with the user query or task requirements. In user-centric settings, relevance directly affects perceived utility and helpfulness of an \ac{llm} output. \\

Retrieval
& When a non-parametric attribution approach is used and its quality depends on retrieval performance
& Retrieval assesses the performance of the retrieval system, which strongly influences downstream attribution quality. \\

\bottomrule
\end{tabular}

\caption{Evaluation guidelines for evidence-based text generation with \acp{llm}. Not all approaches need to be assessed across all evaluation dimensions. We distinguish between core evaluation dimensions, which should be evaluated, and contextual evaluation dimensions, whose applicability depends on the task and system design.}
\label{tab:evaluation_recommendation}
\end{table*}

\textbf{Preservation.} Introduced by \citet{gao-etal-2023-rarr} in the context of post-retrieval attribution, preservation evaluates how much a revised output \( y \) retains from the original \ac{llm}-generated text \( x \) after incorporating retrieved evidence. The evaluation focuses on limiting unnecessary changes during revision. We identify two frequently used preservation metrics. \textit{Preservation Levenshtein}~\cite{gao-etal-2023-rarr} is a lexical overlap metric that penalizes character-level edits between \( x \) and \( y \). \textit{F1-AP}~\cite{gao-etal-2023-rarr} combines attribution and preservation by computing a harmonic mean that incorporates AIS-sentence with preservation signals.

\textbf{Relevance.} This dimension evaluates how well an \ac{llm}-generated text aligns with the user query or task, commonly associated with utility or helpfulness. Relevance is frequently assessed using \ac{llm}-as-a-judge approaches, such as \textit{G-Eval Relevance}~\cite{liu-etal-2023-g}. In addition, the RAGAS framework~\cite{es-etal-2024-ragas} introduces relevance-focused metrics. \textit{RAGAS Answer-Relevancy} computes the semantic similarity between the input question and questions automatically generated from the answer, while \textit{RAGAS Context-Recall} measures how well the retrieved context reflects the ground-truth answer, focusing on whether key information is retrieved and incorporated.

\textbf{Retrieval.} In non-parametric attribution settings, retrieval assesses whether relevant evidence is provided to the \ac{llm}. As expected, evaluation relies on retrieval-based metrics. Frequent metrics include \textit{Precision@k} and \textit{Recall@k}, which assess relevance and coverage of retrieved documents~\cite{ramu-etal-2024-enhancing}, as well as \textit{Mean Reciprocal Rank}, which captures the rank position of the first relevant item~\cite{xiao2025quillquotationgenerationenhancement}. In addition, \textit{RAGAS Context-Entity-Recall} evaluates entity-level recall by computing the fraction of ground-truth entities present in the retrieved context~\cite{es-etal-2024-ragas}.

\subsection{Evaluation Guidelines}
\label{sec:evaluation_guidelines}

In Table~\ref{tab:evaluation_recommendation}, we outline guidelines for evaluating evidence-based text generation with \acp{llm}. Although we identify seven evaluation dimensions, their relevance varies across approaches. We define \textit{core} evaluation dimensions, namely attribution or citation, which depend on evidence availability, and correctness, which should always be evaluated. In addition, \textit{contextual} dimensions include linguistic quality, preservation, relevance, and retrieval, whose applicability depends on the task, system design, and application setting. This distinction ensures that core aspects of evidence-based text generation with \acp{llm} are consistently assessed, promoting more standardized evaluation practices.

For the core evaluation dimensions, we provide a comparative overview of frequently reused evaluation metrics in Table~\ref{tab:metric-comparison} in the appendix, which summarizes their measured aspects, applicability, and limitations to support informed metric selection. Additional evaluation metrics are discussed in Appendix~\ref{sec:additional_metrics}, while task-specific analyses and trends of evaluation dimensions and methods are presented in Appendix~\ref{sec:appendix_evaluation_methods} and Appendix~\ref{sec:appendix_evaluation_dimensions}.

\begin{takeawaybox}
\textbf{Takeaways.} Evaluation of evidence-based text generation with \acp{llm} is inherently multidimensional, yet current practices often assess dimensions in isolation. While not every approach requires evaluation along all dimensions, their interactions are critical for meaningful assessment, and evaluation choices implicitly prioritize certain \ac{llm} behaviors. For correctness in long-form text generation, there is a fundamental trade-off between scalability and semantic coverage. Human evaluation remains dominant due to the difficulty of capturing nuanced factual errors. Automated metrics are more scalable but capture only partial signals: lexical and inference-based methods rely on surface or entailment cues, while \ac{llm}-as-a-judge approaches extend coverage but require careful alignment with human judgments. Consequently, current automated correctness metrics provide only indicative signals, leaving substantial room for future research.
\end{takeawaybox}
\section{Discussion and Future Directions}
\label{sec:discussion}

Evidence-based text generation with \acp{llm} has emerged as a rapidly growing research area, with 75\% of surveyed works published after 2023. Despite this progress, key limitations identified by \citet{huang-chang-2024-citation} remain largely unresolved. Building on these insights, we synthesize our findings to highlight four central limitations that define important directions for future research.

\textbf{Parametric and Hybrid Attribution.} Parametric attribution remains an open and underexplored challenge, with existing model- and data-centric approaches facing significant limitations in scalability and generalizability. Hybrid attribution, which combines parametric and non-parametric signals, offers a pragmatic pathway to improve attribution even when parametric methods remain imperfect. By allowing partial or coarse parametric signals to complement retrieved evidence, hybrid approaches can help surface the boundaries of model-internal knowledge, mitigate retrieval limitations, and provide richer attribution signals than either approach alone. At the same time, advances in parametric attribution remain critical for strengthening hybrid methods, suggesting a co-evolution rather than a strict sequential dependency between the two.

\textbf{Evaluation Standards.} Despite 300 identified evaluation metrics, only two frameworks and two benchmarks are frequently reused among studies. This emphasizes the urgent need for standardized evaluation frameworks to enable fair and consistent comparison across methods. Future work should ensure these frameworks are adaptable to the diverse tasks within evidence-based text generation with \acp{llm} and comprehensively cover all seven evaluation dimensions outlined in this survey. Additionally, the wide variation in human evaluation, with many studies introducing unique and non-standardized metrics, underscores the need for automated and scalable evaluation approaches.

\textbf{Explainable Citations.} The citation behavior of \acp{llm} remains underexplored. These models may exhibit citation-related biases similar to those seen in human authorship. To improve explainability of \ac{llm}-generated texts, users should understand why \acp{llm} select a source from multiple candidates. Transparent citation reasoning is a prerequisite to identify biases and increase trust. Future work should systematically analyze the citation behavior of current approaches and improve the explainability of citation reasoning.

\textbf{Multimodal Evidence.} Evidence-based text generation with \acp{llm} remains text-centric, with text constituting the citation modality in 96\% of studies, indicating that multimodal evidence is largely underexplored. A key open challenge is combining evidence across modalities and evidence levels, such as integrating visual evidence, tables, and graphs within a single response. This requires determining how heterogeneous evidence supports shared claims, and how it should be selected, weighted, and cited in \acp{llm} outputs.
\section{Conclusion}
\label{sec:conclusion}

We surveyed 134 papers on evidence‑based text generation with \acp{llm}, introduced a novel taxonomy, categorized 300 evaluation metrics into seven dimensions and six methods and extracted 231 unique datasets. The field is still fragmented with varied terminology and evaluation standards. Our synthesis offers a clear reference point for advancing the reliability and verifiability of \acp{llm}.

\section*{Limitations}
Even though we followed a systematic study design, including a thorough keyword-based literature search, careful inclusion classification, and iterative refinement of paper annotation, certain limitations apply to our study. The restriction to a single primary search string may have led to the omission of some relevant studies. To assess this risk, we conducted a sensitivity analysis using an expanded set of adjacent search terms, which identified only a small additional proportion of relevant studies (4\%). This indicates that the risk from using a single primary search string is small, and that the original search strategy achieved strong coverage within the major sources considered. Maintaining a manageable number of retrieved studies remains indispensable for manual inclusion screening, particularly since current \ac{llm}-based approaches remain limited compared to human expertise. To further mitigate incomplete literature coverage, we selected nine widely used databases that collectively cover a substantial portion of the examined field. We also tested multiple search strings against an initial literature sample to optimize retrieval completeness while minimizing overall retrieval volume. For more details, see Appendix \ref{sec:methodology}.

The inclusion screening and categorization of relevant studies inherently involve some degree of subjectivity, as both rely on the judgments of the researchers. While each decision was informed by the authors’ domain expertise, this process may still introduce bias. To mitigate these limitations, we established explicit inclusion criteria, resolved disagreements through discussion until consensus was reached, and measured inter-annotator agreement to ensure consistent application of the annotation schema. Additionally, the annotation schema was regularly reviewed and refined to maintain a shared and consistent understanding of each category.

Finally, certain coverage biases may remain. Our restriction to English-language studies may underrepresent non-English research and regional venues, although English is the dominant language in this field. Moreover, focusing on publicly accessible sources may underrepresent non-public or industry-internal work. Including arXiv helped capture recent developments and reflects common dissemination practices in rapidly evolving research areas, but does not address the limited visibility of proprietary research. These factors should be considered when interpreting our findings.

\section*{Acknowledgments}
We thank AFM Mohimenul Joaa and Kyuri Im for their efforts during our data collection.

The authors acknowledge the financial support by the Federal Ministry of Research, Technology and Space of Germany (BMFTR) and by Sächsische Staatsministerium für Wissenschaft, Kultur und Tourismus in the programme Center of Excellence for AI-research „Center for Scalable Data Analytics and Artificial Intelligence Dresden/Leipzig“, project identification number: ScaDS.AI.

Tobias Schreieder is supported by the BMFTR through a Software Campus project, project identification number: 16|S23070.

Tim Schopf is supported by a scholarship of the German Academic Exchange Service (DAAD).

We used AI-based assistance tools to support language editing, minor formatting, and coding tasks. These tools did not contribute to the intellectual content or scientific conclusions. All content was reviewed by the authors, who assume full responsibility for the publication.

\bibliography{anthology,custom}

\appendix
\section{Data Availability}
\label{sec:availability}

Our dataset, which includes all 134 annotated publications, 300 extracted evaluation metrics, and 231 datasets, is publicly available under the BSD 3-Clause license at \url{https://github.com/faerber-lab/AttributeCiteQuote}.

This dataset facilitates the reproduction of the experiments and analyses described in this study and serves as a foundation for future research.

\section{Methodology}
\label{sec:methodology}

\begin{figure}[ht!]
    \centering
    \includegraphics[width=\linewidth]{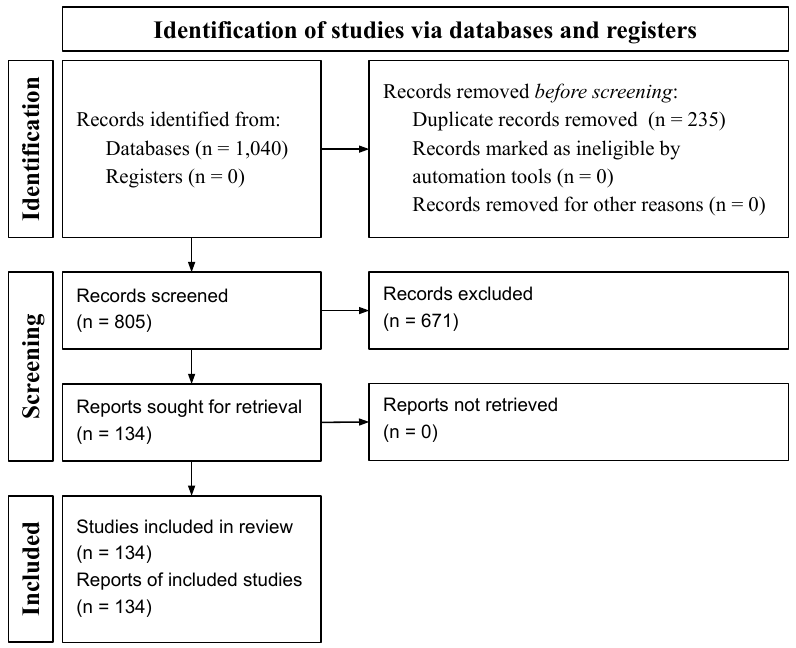}
    \caption{Systematic literature review process following the PRISMA protocol.}
    \label{fig:prisma}
\end{figure}

To provide a comprehensive overview of the research landscape, we conducted a systematic mapping study following the guidelines outlined by \citet{petersen2008systematic}. To ensure transparency and reproducibility in the literature review process, we report the identification, screening, and inclusion of studies using a PRISMA 2020 flow diagram~\citep{Page2021Prisma}, shown in Figure~\ref{fig:prisma}. The main steps are detailed in the following subsections.

\subsection{Literature Search and Inclusion Criteria}
\label{sec:literature_search}

We identified targeted keywords to guide the literature search on evidence-based text generation with \acp{llm}. To stay focused on the \ac{llm}-based paradigm while covering related concepts, we applied the following search string:

\textit{(``\acl{llm}'' OR ``llm'') AND \\ (``citation'' OR ``attribution'' OR ``quote'')}

To identify the most appropriate keywords, we compiled an initial set of literature, including, but not limited to, the position paper by \citet{huang-chang-2024-citation} and the survey papers by \citet{li2023surveylargelanguagemodels} and \citet{zhang2023largelanguagemodelsmeet}. Using this initial literature, we systematically evaluated 15 search strings to identify those offering both high recall (i.e., the proportion of papers from the initial list successfully retrieved) and a limited result set size (i.e., the number of retrieved papers). Controlling the result set size was critical to ensure that manual screening of each candidate paper remained feasible. We deliberately decided against using \ac{llm}-based screening, which would have been necessary with a very high result set size, to ensure the quality and reliability of our annotations and the overall survey. Even minor changes to the search terms, such as applying stemming to the keywords ``citation'', ``attribution'' and ``quote'', led to a 1,207\% increase in retrieved papers (exceeding 10,000 publications) with negligible recall improvement, and were therefore deemed unsuitable. Similarly, adding terms such as ``reference'', ``source'', or ``evidence'' did not yield further recall gains. Consequently, we adopted the keyword configuration above that achieved the optimal balance between recall and result set size.

The final search string was matched against the title and abstract of each publication. To ensure comprehensive coverage, we queried nine literature databases, following \citet{schneider-etal-2022-decade}. The ACL Anthology is a primary repository for leading conferences and journals in \ac{nlp}. The ACM Digital Library, IEEE Xplore, ScienceDirect, and Springer Nature cover key venues across the broader computer science domain. We also included arXiv to capture recent publications not yet in the queried databases, as well as relevant proceedings from major machine learning conferences.

In February 2025, we ran the search and initially identified 1,040 publications across all queried databases. After removing duplicate records, 805 unique publications remained and were used for screening. Subsequently, we used the following inclusion criteria to identify studies relevant to our research focus: (1) studies that aim to generate natural language text with \acp{llm}, (2) studies that deliberately incorporate references to sources of evidence during text generation, and (3) studies that are written in English with the full texts electronically accessible. The accessibility requirement served to facilitate full-text screening and did not lead to the exclusion of any otherwise eligible publications. Publications not meeting all criteria were excluded from the final dataset.
The first two authors screened all titles and abstracts, consulting full texts as needed. The screening workload was divided between the annotators, each reviewing a subset of publications. Both annotators are domain experts in computer science with relevant research experience. Each annotator flagged uncertain cases, which were subsequently discussed jointly until consensus was reached. This process resulted in 134 included publications, as shown in Table~\ref{tab:e-databases}.

\begin{table}[t]
    \centering
    \small
    \begin{tabular}{l c}
    \toprule
    \textbf{Literature Database} & \textbf{No. of Papers} \\
    \midrule
    ACL Anthology & 54 \\
    ACM Digital Library & 7 \\
    arXiv & 59 \\
    ICML Proceedings & 0 \\
    ICLR Proceedings & 3 \\
    IEEE Xplore & 4 \\
    NeurIPS Proceedings & 3 \\
    ScienceDirect & 0 \\
    Springer Nature & 4 \\
    \midrule
    \textbf{Total} & \textbf{134} \\
    \bottomrule
    \end{tabular}
    \caption{Overview of the queried literature databases and the number of studies included.}
    \label{tab:e-databases}
\end{table}

\textbf{Inter-Annotator Agreement.} To assess the reliability of the inclusion screening, both annotators independently labeled a random sample of 50 studies prior to discussion. Inter-annotator agreement was measured using Krippendorff’s $\alpha$ and yielded a value of 1.0, indicating perfect agreement.

\textbf{Sensitivity Analysis.} To assess the robustness of our literature search, we conducted a sensitivity analysis using an expanded set of adjacent terms commonly used in the field:

\textit{(``llm'' OR ``language model'') AND (evidence-based'' OR evidence linking'' OR grounded generation'' OR source-grounded'' OR provenance tracing'' OR source linking'' OR verifiable generation'')}

The query was applied to the ACL Anthology and arXiv, which together cover 84\% of the studies included in our review. The expanded search retrieved 126 studies. After deduplication with the initial literature corpus, 103 unique papers remained and were screened according to our protocol. This process identified five additional relevant studies. Overall, the results indicate that the original search strategy already captured the vast majority of relevant literature within these major sources, with the sensitivity search yielding only a small additional fraction of relevant studies (4\%). The additional studies did not affect the structure of the proposed taxonomy. The five relevant but non-critical studies identified through this process are listed here for transparency \cite{schimanski-etal-2024-towards, hsu-etal-2024-calm, li-etal-2024-llatrieval, yue-etal-2024-evidence, hennigen2024towards}. These were not incorporated into the main corpus, as their inclusion did not affect the structure of the proposed taxonomy.

\begin{table*}[ht!]
    \centering
    \small
    \begin{tabular}{l p{12cm}}
    \toprule
    \textbf{Contribution Type} & \textbf{Description} \\
    \midrule
    Approach & An approach consists of a set of novel methods, techniques, and procedures that need to be systematically executed to achieve a concrete goal. \\
    
    Application & An application is a documented implementation of an existing approach, technique, or method in the form of a software library, prototype, or full application system. \\
    
    Resource & A resource is a published dataset that supports approaches, techniques, methods, or applications, e.g., text corpora or benchmarks. \\
    
    Evaluation & Evaluations of existing approaches, techniques, or methods as well as the introduction of new evaluation approaches including, e.g., new metrics or frameworks. \\
    
    Survey & A survey analyses and synthesizes findings from multiple studies to systematically review a research field or gather evidence on a topic. \\
    
    Position & A position paper presents a personal perspective on the suitability or direction of a specific research aspect, without presenting new empirical evidence. \\
    \bottomrule
    \end{tabular}
    \caption{Categorization scheme for contribution types adapted from \citet{schneider-etal-2022-decade}.}
    \label{tab:contribution-type-scheme}
\end{table*}

\subsection{Categorization Scheme and Method}
\label{sec:categorization_scheme}

We categorized each publication by contribution type: approach, application, resource, evaluation, survey, and position. The detailed categorization scheme, adapted from \citet{schneider-etal-2022-decade}, is presented in Table~\ref{tab:contribution-type-scheme}. Other dimensions are based on \citet{huang-chang-2024-citation} and \citet{li2023surveylargelanguagemodels}, and were iteratively refined following the methodology of \citet{petersen2008systematic}. All 134 studies were categorized along the defined dimensions, with annotation split between the first two authors. Regular calibration rounds refined the scheme and resolved labeling ambiguities. Each paper could be assigned multiple values per dimension if needed. Based on this categorization scheme, we derived a multidimensional taxonomy for evidence-based text generation with \acp{llm}, which organizes the annotated studies along conceptually distinct facets. In addition, we annotated the publication date of each study, using the earliest available version such as preprints. This ensures that emerging trends can be analyzed as they appear, without the delay introduced by formal publication timelines.

\subsection{Design of the Taxonomy}
\label{sec:taxonomy}

We adopt a multidimensional taxonomy based on a facetted classification approach, following the principles outlined by \citet{Crowston2004FacettedClassification}. In a facetted design, independent conceptual dimensions describe different aspects of the study, enabling flexible representation of complex approaches without enforcing artificial mutual exclusivity. This approach is well suited for evidence-based text generation with \acp{llm}, where methods often combine multiple mechanisms that cannot be captured by a single hierarchical structure.

Our taxonomy consists of three dimensions that together characterize papers on evidence-based text generation with \acp{llm}: (1) attribution approach, (2) citation characteristics, and (3) task. These dimensions represent distinct analytical perspectives on approach design choices. Since real-world approaches frequently combine mechanisms, the taxonomy allows multi-label assignments within a dimension, while each dimension captures a separate facet of the approach.

The \textbf{attribution approach} dimension describes how generated text is linked to supporting evidence. Within this dimension, parametric and non-parametric attribution form subdimensions, and approaches are further organized into finer-grained classes that capture different ways of tracing evidence back to \ac{llm}-generated text. 
The \textbf{citation characteristics} dimension describes what evidence is made available to users and how it is presented. This includes the modality, evidence level, style, visibility, and frequency of citations.
The \textbf{task} dimension specifies the functional goal of the approach, such as \ac{qa}, grounded text generation, summarization, related work generation, citation text generation, and fact verification.

By combining these dimensions, the multidimensional taxonomy supports fine-grained analysis of heterogeneous approaches and enables consistent comparison of approaches while acknowledging the multifaceted nature of evidence-based text generation with \acp{llm}. The taxonomy in Figure~\ref{fig:taxonomy} and the annotated dataset of 134 papers in Table~\ref{tab:classification_papers} can be read from left to right, where each paper is categorized across all dimensions, with multi-label annotations when a paper spans multiple classes.

\begin{table}[t!]
    \centering
    \small
    \begin{tabular}{l c}
    \toprule
    \textbf{Taxonomy Dimension} & \textbf{Krippendorff’s $\alpha$} \\
    \midrule
    Contribution Type & 0.73 \\
    Non-Parametric & 0.82 \\
    Citation Modality & 0.79 \\
    Evidence Level & 0.83 \\
    Citation Style & 0.77 \\
    Citation Visibility & 1.00 \\
    Citation Frequency & 0.92 \\
    Task & 0.88 \\
    \bottomrule
    \end{tabular}
    \caption{Inter-annotator agreement for literature annotation measured using Krippendorff’s $\alpha$. The scores are macro-averaged across labels within each dimension.}
    \label{tab:iaa-taxonomy}
\end{table}

\textbf{Inter-Annotator Agreement.} To assess the reliability of the taxonomy annotation, both annotators independently labeled a random subset of 30 publications. Inter-annotator agreement was measured using Krippendorff’s $\alpha$. Given the multidimensional and multi-label nature of the taxonomy, agreement was computed separately for each taxonomy dimension. Within each dimension, $\alpha$ was calculated for individual labels that appeared at least five times in the annotated subset to ensure stable estimates. For each dimension, we report the macro-average across its labels.

The agreement scores are shown in Table~\ref{tab:iaa-taxonomy}. No studies employing parametric attribution approaches were present in the annotated sample, so agreement could not be computed for this dimension. The observed values indicate acceptable to strong agreement across all taxonomy dimensions, according to commonly used interpretation guidelines for Krippendorff’s $\alpha$. The comparatively lower agreement for contribution type is consistent with its multi-label nature.

\section{Research Landscape}
\label{sec:research_landscape}

\begin{figure}[t]
    \centering
    \includegraphics[width=\linewidth]{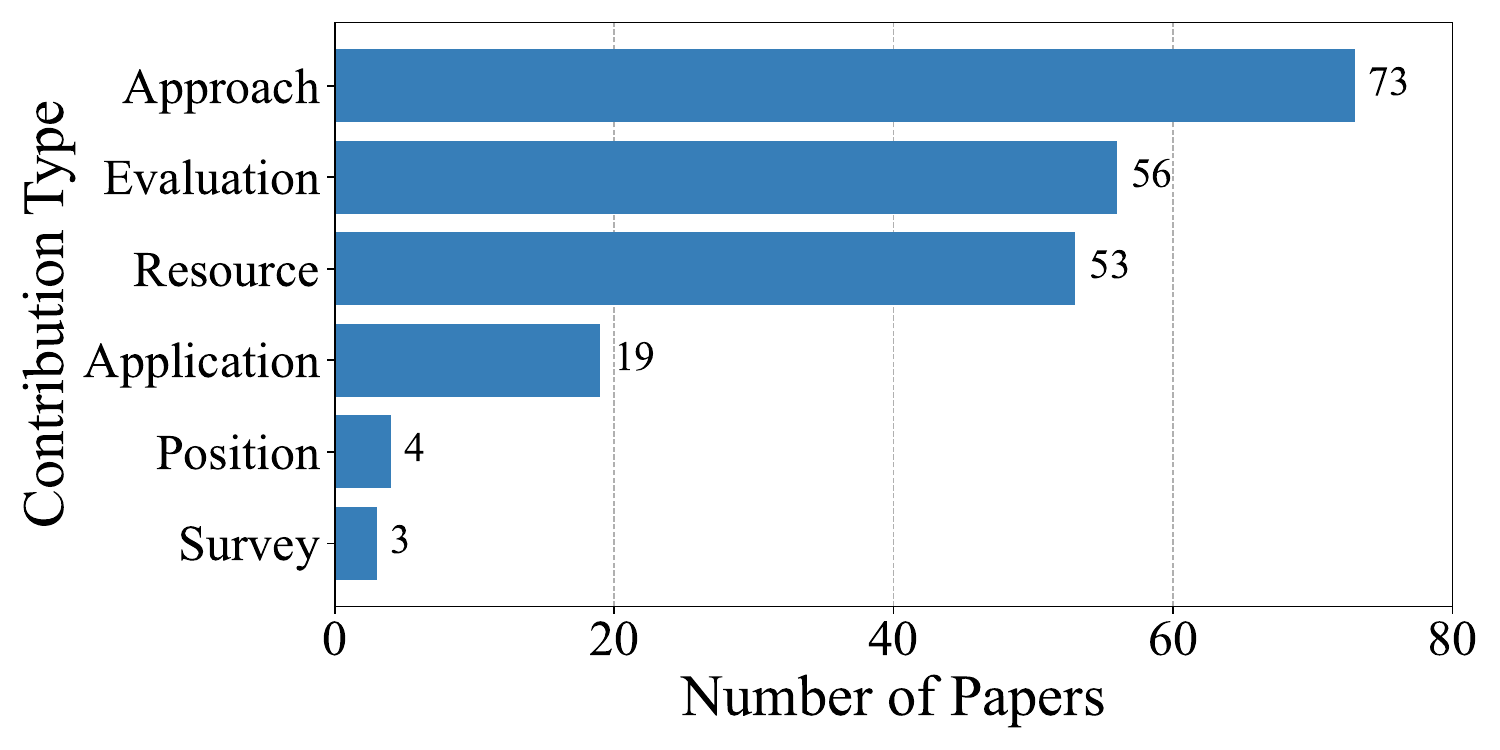}
    \caption{Number of studies by contribution type. Each paper can be assigned to multiple contribution types, so totals exceed 134.}
    \label{fig:contribution_types}
\end{figure}

Figure~\ref{fig:contribution_types} shows the distribution of publications by contribution type. We observe that most studies propose novel approaches for evidence-based text generation with \acp{llm}. A substantial number also introduce new resources and focus on evaluation, underscoring the growing attention these aspects receive within the community. Further, this highlights the necessity of not only reviewing methodological contributions but also systematically mapping existing evaluation approaches and resources, a gap this survey addresses in Section~\ref{sec:evaluation} and Appendix~\ref{sec:evaluation_resources}. Application, position, and survey studies are relatively underrepresented.

\section{Extended Analysis and Trends of Evidence-based Text Generation}
\label{sec:additional_results}

This section presents an extended analysis of evidence-based text generation with \acp{llm}, complementing the analyses in Section~\ref{sec:evidence_based_text_generation}. In particular, we highlight representative works associated with each dimension of the proposed taxonomy, offer task-specific analyses, and discuss trends that have emerged over time.

\subsection{Attribution Approach}
\label{sec:additional_attribution_approach}

Attribution refers to the process of tracing \ac{llm}-generated text back to supporting evidence. In our taxonomy, we distinguish between parametric and non-parametric attribution approaches. Non-parametric attribution is applied substantially more often, with 126 studies, compared to 25 studies on parametric attribution. Within parametric attribution, 18 studies evaluate the ability of pure \acp{llm} to correctly attribute a generated text to a source of evidence (e.g., by generating citations), while four studies address data-centric attribution and three studies focus on model-centric attribution. For non-parametric attribution, where external evidence sources are incorporated into the \ac{llm}, the majority of approaches adopt post-retrieval attribution, with 73 studies. In-context attribution is used in 25 studies. Post-generation attribution is explored in 22 studies, whereas in-generation attribution is comparatively rare, with only five studies.

\subsubsection{Parametric Attribution}
\label{sec:additional_parametric_attribution}

\textbf{Pure \acp{llm}} rely solely on the model’s inherent attribution behavior. Studies in this category focus on evaluating the attribution behavior of \acp{llm}, for example, by assessing the presence and quality of citations generated by the model~\cite{byun-etal-2024-reference, malaviya-etal-2024-expertqa}. Early work highlights substantial limitations. \citet{Pride2023CORE-GPT} report that citations generated by GPT-3.5 are factually correct only 22\% of the time, and that GPT-4 does not improve this rate (20\%). Similarly, \citet{Zuccon2023ChatGPTCitation} find that while GPT-3.5 generates correct or partially correct answers in 51\% of cases, its accompanying references correspond to real sources only 14\% of the time. While pure \acp{llm} are typically prompted to provide citations, \citet{Moayeri2024CitationHallucination} observed that GPT models occasionally generate citations spontaneously.

\textbf{Model-Centric} attribution adjust the model architecture or training objectives. In our review, we found that two papers with the contribution type ``approach'' exemplify this strategy.
First, \citet{chu-etal-2025-towards} propose FARD, which trains a student model to imitate citation-grounded rationales distilled from a teacher model, replacing chain-of-thought reasoning to improve causal-aware attribution. Second, \citet{Khalifa2024SourceAwareTraining} introduce source-aware training, where the \ac{llm} learns to associate knowledge with document identifiers during pretraining, followed by instruction tuning that encourages explicit citations.

\textbf{Data-Centric} attribution curates, augments, or synthetically generates data. We identified three papers with the contribution type ``approach'' that fall under this category. \citet{li-etal-2024-improving-attributed} introduce an automatic preference optimization framework that models attribution as a preference learning task using both curated and automatically synthesized citation preference data. \citet{huang-etal-2024-learning} propose FRONT, a training approach that supervises \acp{llm} with fine-grained supporting quotes to guide citation generation. \citet{lu-etal-2025-wasa} frame attribution as a watermarking task, enabling \acp{llm} to embed source-identifying signals into the text.

\begin{figure*}[ht!]
    \centering
    \includegraphics[width=\linewidth]{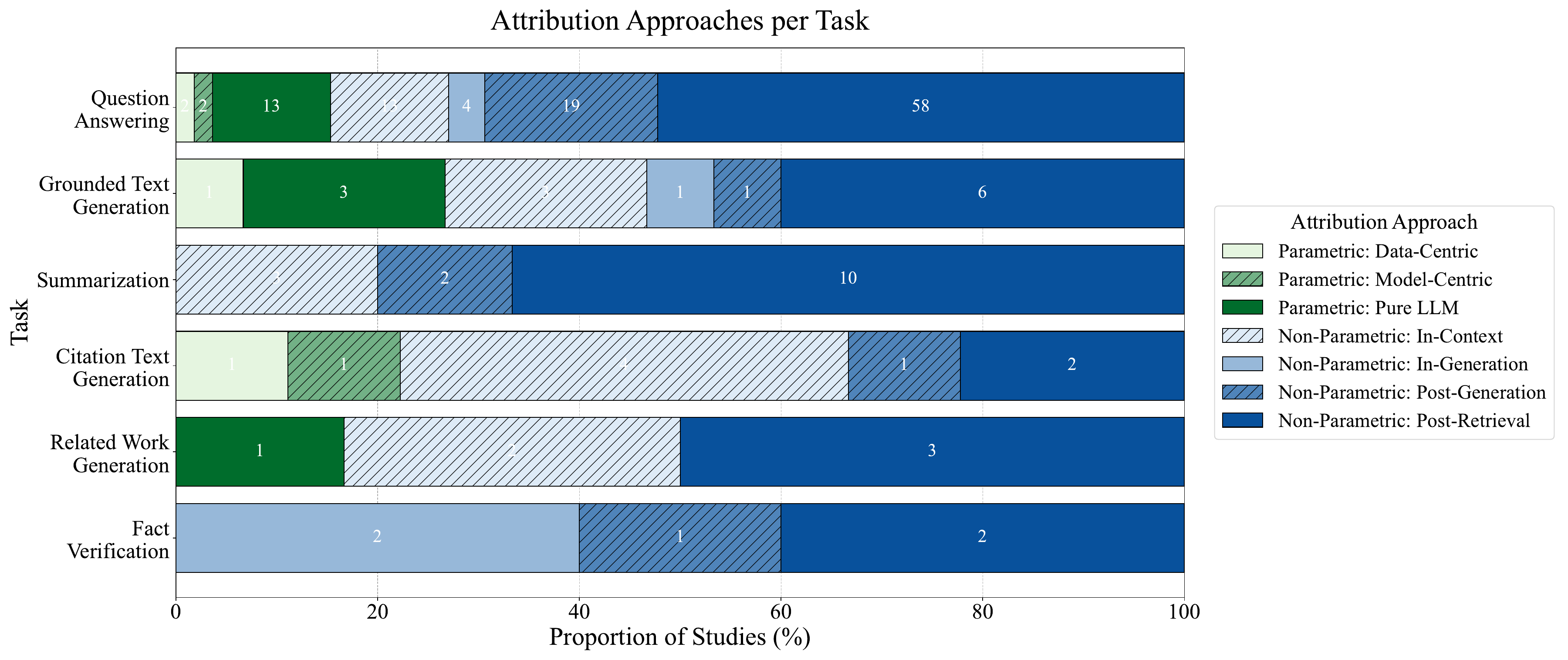}
    \caption{Distribution of attribution approaches across tasks in evidence-based text generation with \acp{llm}. The bars represent the relative percentages of parametric attribution approaches (green) and non-parametric attribution approaches (blue) for each task. The absolute number of papers for each attribution approach is reported alongside the bars. Tasks include \ac{qa}, grounded text generation, summarization, citation text generation, related work generation, and fact verification. Each paper can be assigned to multiple tasks, so totals exceed 134.}
    \label{fig:attribution_per_task}
\end{figure*}

\textbf{Task-specific Analysis.} Figure \ref{fig:attribution_per_task} shows how attribution approaches distribute across tasks. Parametric attribution appears only in \ac{qa}, grounded text generation, citation text generation, and related work generation~\cite{Zuccon2023ChatGPTCitation, lu-etal-2025-wasa, huang-chang-2024-citation, byun-etal-2024-reference}. As observed from our annotated dataset, summarization always relies on non-parametric attribution because the task requires one or more input documents to serve as evidence for the summary. Fact verification, in principle, could be performed using only parametric model knowledge, but the lack of parametric attribution in current studies indicates that \acp{llm} are not yet considered sufficiently reliable for this purpose~\cite{buchmann-etal-2024-attribute, asai2024selfrag}. As a result, existing work predominantly relies on non-parametric attribution to ensure factuality. 72\% of parametric attribution approaches evaluate pure \acp{llm} without modifying the model architecture or training data. Model-centric and data-centric attribution is explored only in tasks that are already widely studied and that align closely with the capabilities of state-of-the-art \ac{llm} chatbots~\cite{openai2023gpt4}. This makes them natural targets for developing parametric attribution approaches, as these tasks benefit most directly from models that can generate text attributed to parametric knowledge. The observed patterns indicate that parametric attribution in pure \acp{llm} remains unreliable for tasks with strong factuality requirements~\cite{Pride2023CORE-GPT}, while data-centric and model-centric approaches exist only for a small subset of tasks and still face significant limitations~\cite{huang-chang-2024-citation}. As a result, tasks that demand robust evidence grounding continue to rely predominantly on non-parametric methods, which highlights key limitations of current \ac{llm} architectures.

\textbf{Trends.} Non-parametric attribution approaches appeared substantially earlier than parametric ones, with the first non-parametric studies published in Q1 2022 and the first parametric attribution approach emerging only in Q3 2023. Since then, parametric attribution has shown little momentum: new approaches appear only occasionally and do not form a clear upward trend. This stagnation reflects the technical difficulty of adapting or training \acp{llm} to emit reliable attribution signals, as well as the limited scalability and generalizability of current data-centric and model-centric approaches~\cite{lu-etal-2025-wasa, Khalifa2024SourceAwareTraining}. In contrast, non-parametric attribution continues to grow steadily, indicating that parametric attribution has not yet matured into a widely adopted methodology.

\subsubsection{Non-Parametric Attribution}
\label{sec:additional_non_parametric_attribution}

\textbf{Post-Retrieval} attribution retrieves external information before text generation and uses it to ground the model’s output in evidence. The first post-retrieval approach included in our survey is GopherCite, which is trained to generate answers together with in-line citations~\cite{menick2022teachinglanguagemodelssupport}. Attribution is achieved by training the model to generate verbatim quotes from retrieved documents within a fixed output format, with supervised fine-tuning and reinforcement learning from human feedback encouraging answers that are both correct and explicitly supported by the cited evidence. In contrast to GopherCite’s joint generation of answers and citations, PURR adopts a revision-based perspective: it first generates search queries from an ungrounded given statement to retrieve relevant evidence, which is then used by a small editor model to revise the original statement~\cite{chen2023purrefficientlyeditinglanguage}. PURR can be interpreted as either post-retrieval or post-generation. From a post-generation perspective, it revises an already generated statement, while from a post-retrieval perspective, the attributed output depends on evidence retrieved beforehand. Since we define attribution as the process of tracing generated text back to supporting evidence, we classify PURR as post-retrieval, as the grounding evidence is explicitly retrieved before text generation. \citet{huang-etal-2024-training} focus on improving post-retrieval attribution through fine-grained reward signals applied during training. Within their method, retrieved passages serve as supervision signals, and fine-grained rewards evaluate how well segments of the generated text align with the evidence. In contrast to previous approaches, MIRAGE derives attribution from model internals by estimating how strongly each retrieved context token contributes to each generated answer token~\cite{qi-etal-2024-model}. This enables token-level attribution without modifying the underlying model, as saliency scores are used to trace the generated output back to specific parts of the retrieved evidence.

\textbf{Post-Generation} attribution reverses the order of post-retrieval attribution by first generating an answer using parametric model knowledge and then revising it using retrieved non-parametric evidence. \citet{gao-etal-2023-rarr} introduce a revision step in their system, RARR, which performs posterior verification by assessing the correctness of the \ac{llm} response with reference to the retrieved documents and making adjustments when necessary. \citet{yan2024atomicfactdecompositionhelps} extend this idea with ARE, which decomposes a generated long-form answer into atomic facts and retrieves evidence for each fact individually, enabling more fine-grained verification and edits than the single-step, answer-level revision used in RARR. In contrast, CEG follows a regeneration-based strategy in which evidence is retrieved after an initial answer is generated, and the response is regenerated until all statements are supported by citations \cite{li-etal-2024-citation}. This differs from RARR, which performs a single targeted revision rather than iteratively regenerating the full answer. \citet{huang-etal-2024-advancing} extend post-generation attribution with START, a self-improving framework that begins with synthetic attribution data and then iteratively samples candidate responses, retrieves evidence, and applies fine-grained preference optimization to improve attribution.

\textbf{In-Generation} attribution dynamically determines when additional evidence is needed and activates the retrieval system during the generation process. A prominent example is Self-\ac{rag}~\cite{asai2024selfrag}, which introduces self-reflection tokens enabling the \ac{llm} to adaptively retrieve relevant evidence and iteratively critique and revise its outputs during inference. Other notable approaches include AGREE~\cite{ye-etal-2024-effective}, which trains \acp{llm} to retrieve and cite evidence dynamically during generation, and Think\&Cite~\cite{li-ng-2025-think}, which employs self-guided tree search combined with reward modeling to progressively enhance attribution through iterative evidence gathering. Additionally, NEST~\cite{li2024nearest} performs token-level retrieval at each generation step, incorporating relevant spans into the output.

\textbf{In-Context} attribution does not require retrieval models, as the evidence is directly provided within the prompt.  This setting has been adopted in a variety of scenarios. For instance, several studies use in-context attribution to evaluate their attribution approaches by supplying a curated set of source documents, rather than relying on retrieval systems~\cite{zhang-etal-2025-longcite, cohen-wang2024contextcite, slobodkin-etal-2024-attribute}. Furthermore, some tasks inherently require in-context attribution. A typical example is document-level summarization, where multiple documents are given to the \ac{llm} for attributed summarization. Another example is citation text generation, where the \ac{llm} is prompted to generate citation-worthy text that integrates the content of a given scientific paper into a surrounding context, such as a related work section~\cite{Anand2023KG-CTG, gu-hahnloser-2024-controllable}.

\textbf{Task-specific Analysis.} Figure \ref{fig:attribution_per_task} shows how non-parametric attribution paradigms distribute across the six tasks. Post-retrieval attribution is by far the dominant approach, appearing in every task and accounting for the largest proportion within non-parametric approaches in \ac{qa} (62\%), grounded text generation (55\%), summarization (67\%), and related work generation (60\%). Its prevalence reflects the simplicity and effectiveness of current \ac{rag} pipelines. In contrast, post-generation attribution is adopted less frequently, representing only 20\% of non-parametric approaches and appearing mainly in \ac{qa}. Only a limited set of tasks benefits from workflows that combine parametric and non-parametric knowledge~\cite{gao-etal-2023-rarr, ramu-etal-2024-enhancing}, while most settings rely effectively on post-retrieval attribution with outputs grounded directly in retrieved evidence. Post-generation attribution remains valuable, however, in scenarios where reduced reliance on retrieval is advantageous. For example, in \ac{qa} tasks that require models to provide an output even under incomplete or imperfect evidence, models can first draw on parametric knowledge and then verify or refine the output using external evidence, whereas post-retrieval approaches may return no answer when retrieval fails. In-context attribution exhibits clear task specificity. It is widely used in summarization, citation text generation, and related work generation, tasks where relevant evidence is naturally provided as input by the user~\cite{slobodkin-etal-2024-attribute, gu-hahnloser-2024-controllable, nishimura-etal-2024-toward}. In many evaluation settings, evidence is provided directly rather than retrieved so that the attribution behavior of \acp{llm} can be assessed independently of retrieval performance~\cite{cao-wang-2024-verifiable, lee-etal-2024-ask}. This applies in particular to \ac{qa} and grounded text generation, where in-context attribution is used when retrieval components are removed and the model’s reasoning is evaluated in isolation. In-generation attribution remains the least explored non-parametric paradigm. To date, it has only been applied to \ac{qa} and fact verification~\cite{asai2024selfrag, li-ng-2025-think}, and has been proposed for grounded text generation~\cite{Tilwani2024NeurosymbolicAI}. Overall, non-parametric attribution shows clear strengths and constraints. In particular, its reliability depends on the availability of accurate evidence and the quality of the underlying retrieval model. It also does not provide insight into the model’s reasoning process. Nevertheless, it remains the prevailing approach for evidence-based text generation with \acp{llm}, with post-retrieval attribution dominating the literature.

\textbf{Trends.} Non-parametric attribution has exhibited a clear and steady increase since its first appearance in Q1 2022. Post-retrieval attribution shows the strongest growth trajectory and has rapidly become the dominant paradigm. Although only 14 post-retrieval attribution approaches were published in 2023, this number increased notably to 56 in 2024. This indicates a significant trend toward adopting \ac{rag} as the state-of-the-art architecture for evidence-based text generation with \acp{llm}. Post-generation and in-context attribution follow similar patterns, with initial work emerging in Q4 2022 and the number of studies increasing from 3 in 2023 to 15 in 2024 for each paradigm. This demonstrates consistent growth, although at a smaller scale compared to post-retrieval attribution. In-generation attribution, by contrast, is the most recent paradigm, appearing first in Q4 2023 with only two approaches in 2023 and three in 2024. These early and sparse instances do not yet form a noticeable trend. Overall, the field continues to grow primarily along non-parametric attribution, driven by rapid growth in post-retrieval attribution approaches, while post-generation and in-context attribution show gradual growth and in-generation attribution remains exploratory.

\subsection{Citation Characteristics}
\label{sec:additional_citation_characteristics}

Following the attribution approaches, this section provides a detailed analysis of the five citation characteristics for evidence-based text generation with \acp{llm}: citation modality, evidence level, citation style, citation visibility, and citation frequency.

\subsubsection{Citation Modality}
\label{sec:additional_citation_modality}

The citation modality specifies the type of evidence cited by an approach. We distinguish four modalities: \textbf{texts}, \textbf{graphs}, \textbf{tables}, and \textbf{visuals}. The vast majority of studies rely on textual evidence, with 96\% of the surveyed papers citing unstructured text as their primary source of evidence. Within text-based approaches, the underlying evidence predominantly comes from a small set of well-established domains~\cite{gao-etal-2023-enabling, malaviya-etal-2024-expertqa}. Table~\ref{tab:datasets} shows the most frequently reused datasets per task. As shown in the table, encyclopedic sources, most notably Wikipedia, are the most frequently reused, followed by scientific literature and general web search content. News articles and synthetic datasets appear less often, while domains such as social media, government documents, and health-related texts are only sparsely represented. The predominance of these general-purpose textual domains suggests that current approaches are primarily evaluated on easily accessible evidence sources, leaving the robustness of evidence-based text generation underrepresented for more specialized or regulated domains. In contrast, non-textual modalities are used less frequently. Graph-based evidence appears in only a small subset of studies, with four papers using this modality, and is typically used to represent structured relationships between entities~\cite{he-etal-2025-evaluating-improving, dehghan-etal-2024-ewek}. Tabular evidence is used even more sparsely, appearing in three studies, and is generally applied when factual grounding requires access to structured numerical or categorical data~\cite{Roy2025RAG, mathur-etal-2024-matsa}. Visual evidence, such as images, is explored in only two approaches, reflecting early efforts toward multimodal evidence~\cite{ma-etal-2025-visa, suri-etal-2025-visdom}.

\textbf{Task-specific Analysis.} The distribution of citation modalities varies across tasks but remains strongly dominated by textual evidence. Summarization, citation text generation, related work generation, and fact verification rely exclusively on text-based evidence. \ac{qa} is the only task that spans all identified citation modalities, although non-textual evidence remains clearly underrepresented. Grounded text generation also exhibits limited modality diversity, with only a single approach incorporating tabular evidence. This task-level distribution highlights that modality diversity in evidence-based text generation with \acp{llm} is uneven and largely concentrated in \ac{qa}.

\textbf{Trends.} Textual evidence has been present since the earliest work on evidence-based text generation, with the first paper appearing in Q2 2021. The number of text-based approaches increased substantially from 26 publications in 2023 to 93 in 2024, reinforcing text as the dominant citation modality over time. Non-textual modalities emerge considerably later and remain rare. Graph-based evidence first appears in Q4 2023 and grows modestly from one study in 2023 to four in 2024. Tabular and visual evidence are the most recent modalities, both first introduced in Q4 2024, with three and two studies published in 2024, respectively. Overall, modality-level trends indicate that non-textual citation modalities remain underrepresented but are beginning to attract increasing research interest.

\subsubsection{Evidence Level}
\label{sec:additional_evidence_level}

\begin{figure*}[t!]
    \centering
    \includegraphics[width=\linewidth]{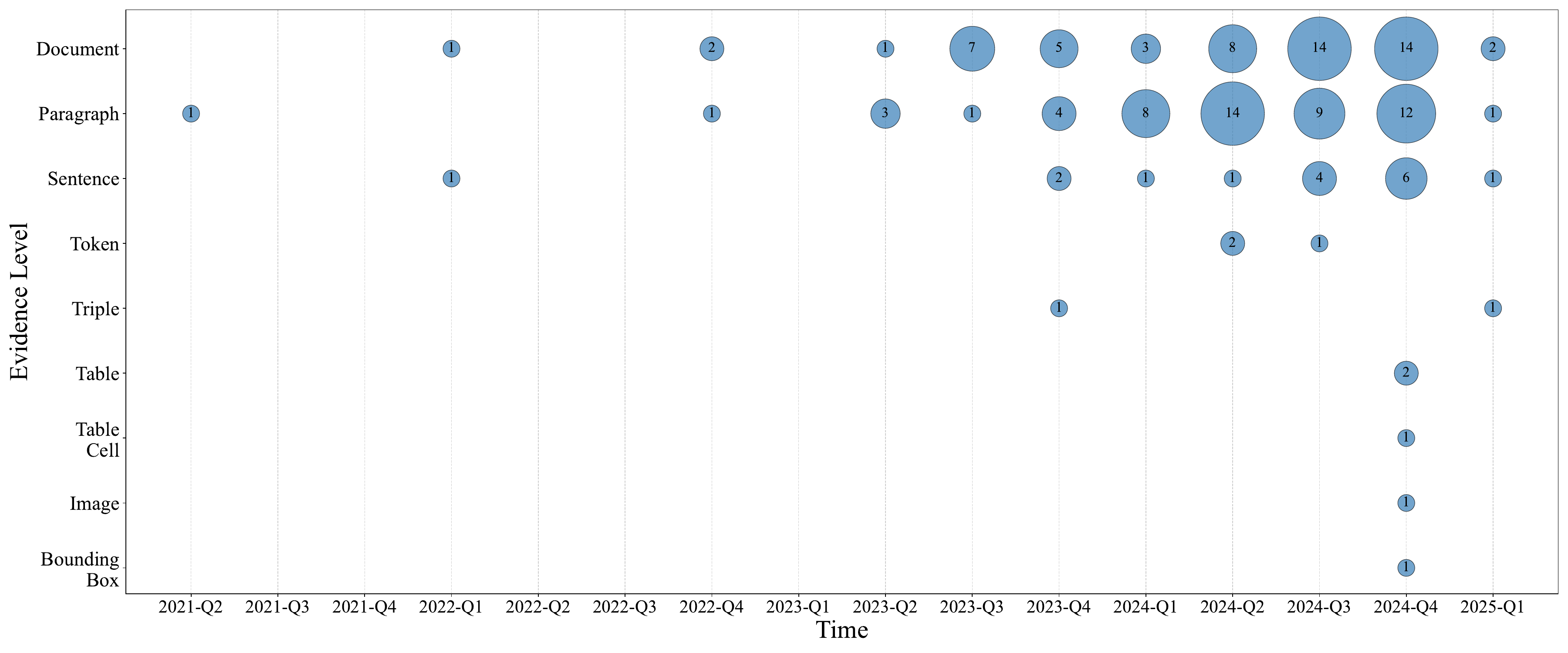}
    \caption{Temporal trends of evidence levels in evidence-based text generation with \acp{llm}. For each evidence level, the figure reports the number of studies adopting that evidence level in each quarter from Q2 2022 to Q1 2025. The size of each bubble corresponds to the number of papers published in the respective quarter using the given evidence level. Q1 2025 includes relevant papers available up to the inclusion deadline in February. Each paper can be assigned to multiple evidence levels, so totals exceed 134.}
    \label{fig:evidence_level_trend_analysis}
\end{figure*}

The evidence level describes the granularity at which cited evidence is linked to \ac{llm}-generated text and is closely related to the underlying citation modality. Rather than specifying the type of evidence, this dimension captures how fine-grained evidence is referenced within a given citation modality.
For textual evidence, multiple evidence levels are observed. Evidence can be cited at the level of a full \textbf{document}, such as an entire scientific article~\cite{Anand2023CitationTextGeneration}, or at a finer granularity such as a \textbf{paragraph}, often corresponding to a retrieved text chunk~\cite{gao-etal-2023-enabling}. Some approaches operate at the \textbf{sentence} level~\cite{xu-etal-2025-aliice}, while a small number attempt attribution at the level of individual \textbf{tokens}~\cite{phukan-etal-2024-peering}. Across all reviewed studies, document-level evidence is used in 43\% of papers and paragraph-level evidence in 40\%, making these the most common choices. Sentence-level attribution appears in 12\% of studies, whereas token-level attribution remains rare at 2\%.
For non-textual modalities, the evidence level is largely determined by the structure of the citation modality itself. Graph-based approaches typically cite evidence at the level of individual \textbf{triples}~\cite{li-etal-2024-towards-verifiable}, while no surveyed study cites an entire graph as a single evidence unit. Tabular evidence is referenced either at the level of a full \textbf{table} or at the level of individual \textbf{table cells}~\cite{mathur-etal-2024-matsa}. Visual evidence is cited either as a complete \textbf{image} or as a specific region, such as a \textbf{bounding box}~\cite{ma-etal-2025-visa}. Due to the limited number of studies employing non-textual modalities, differences in evidence level within these modalities are comparatively small.

\textbf{Task-specific Analysis.} Given the strong dominance of textual citation modalities, task-specific differences in evidence level are discussed primarily for text-based evidence. For non-textual citation modalities, the limited number of studies prevents meaningful task-specific differentiation. \ac{qa} exhibits the widest range of textual evidence levels, spanning from coarse-grained document- and paragraph-level sources to fine-grained sentence- and token-level sources. This reflects the need to support both broad contextual grounding and precise factual claims. Grounded text generation also employs multiple textual evidence levels but predominantly relies on document- and paragraph-level evidence, with finer-grained attribution appearing less frequently. In contrast, citation text generation and related work generation rely on coarse-grained textual evidence, most often citing entire documents. This aligns with conventions in scientific writing, where citations reference complete papers rather than individual passages or sentences. Summarization occupies an intermediate position, with different approaches adopting different textual evidence levels. Most rely on document- or paragraph-level grounding, while a smaller number of studies explore sentence- or token-level evidence to support more fine-grained alignment between source content and generated summaries. Fact verification relies mainly on paragraph- and sentence-level textual evidence, reflecting its emphasis on localized factual support.

\textbf{Trends.} Evidence levels at coarse granularity appear earliest and dominate the literature over time. Paragraph-level evidence is the earliest to emerge, with the first studies appearing in Q2 2021, followed by document-level evidence in Q1 2022. Both show substantial growth, with document-level approaches increasing from 13 publications in 2023 to 39 in 2024, and paragraph-level approaches rising from 8 to 43 over the same period. Sentence-level evidence emerges later and remains less common, growing from two studies in 2023 to twelve in 2024. Token-level evidence remains rare. It appears only in Q2 2024 and is explored by a small number of studies. Evidence levels associated with non-textual modalities, such as triples, tables, table cells, images, and bounding boxes, appear only from late 2023 or 2024 onward and remain sparsely represented. Overall, trends indicate strong growth in coarse-grained evidence levels, with finer-grained and non-textual evidence levels emerging only recently and at a smaller scale.

\subsubsection{Citation Style}
\label{sec:additional_citation_style}

\begin{figure}[ht!]
    \centering
    \includegraphics[width=\linewidth]{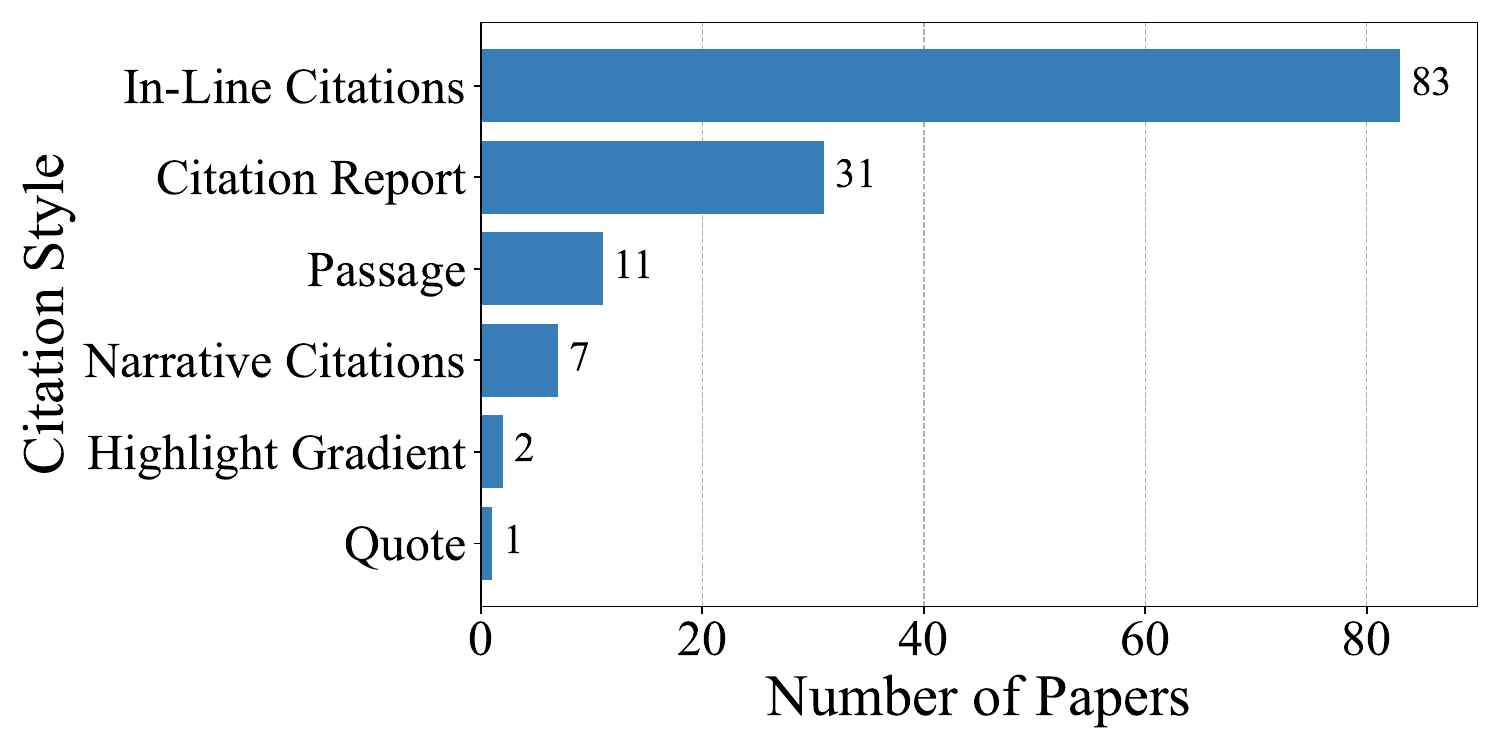}
    \caption{Number of studies using each citation style. Each paper can be assigned to multiple citation styles, so totals exceed 134.}
    \label{fig:citation_style_numbers}
\end{figure}

\begin{figure}[t!]
    \centering
    \includegraphics[width=\linewidth]{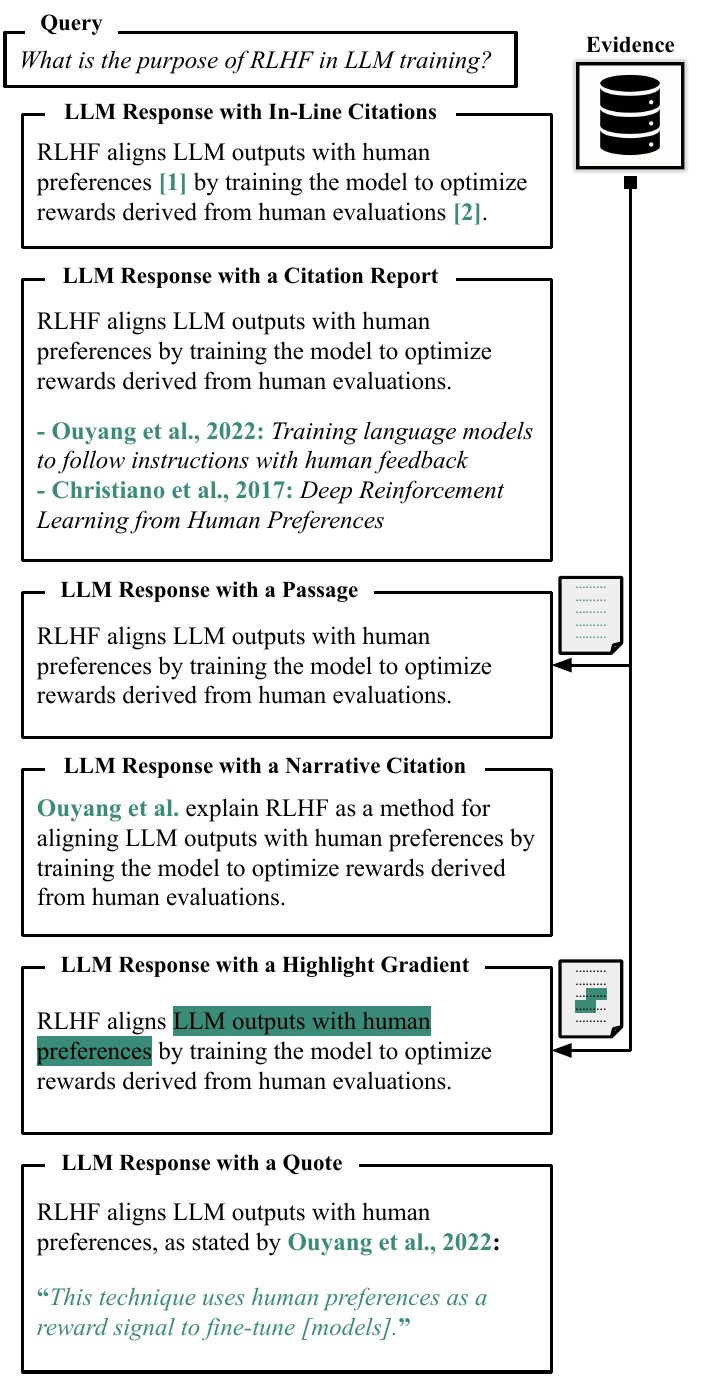}
    \caption{Visualization of different citation styles. For a given query, the \ac{llm} generates a response with citations presented in different styles, enabling users to verify \ac{llm}-generated text against supporting evidence.}
    \label{fig:citation_styles}
\end{figure}

The citation style determines how evidence for \ac{llm}-generated text is presented to the user. We identify six distinct citation styles: in-line citations, citation reports, passages, narrative citations, highlight gradients, and quotes. Figure~\ref{fig:citation_style_numbers} shows the distribution of citation styles across the reviewed studies, highlighting that 62\% of studies rely on in-line citations.
As illustrated in Figure~\ref{fig:citation_styles}, \textbf{in-line citations} are inserted directly after a citation-worthy claim (e.g.,~[1][2])~\cite{huang-etal-2024-training}, enabling users to trace individual statements back to their supporting evidence. Another frequently used style is the \textbf{citation report}, which provides a separate list of references alongside the \ac{llm}-generated output~\cite{bohnet2023attributedquestionansweringevaluation}. While citations are still presented in textual form, users must manually associate reported references with specific generated claims, which makes citation reports less transparent.
Several approaches display only the retrieved \textbf{passage} used by the \ac{llm} during generation, particularly in evaluation settings for attribution approaches~\cite{muller-etal-2023-evaluating}. As shown in Figure~\ref{fig:citation_styles}, the passage from the evidence corpus is presented alongside the \ac{llm}-generated output without any textual or visual citation markers that explicitly link the evidence to specific claims. This requires users to perform the alignment implicitly and makes this citation style less suitable for transparent or user-facing evidence-based text generation with \acp{llm}, while still being useful for methodological evaluation.
\textbf{Narrative citations} integrate references into the natural flow of the generated text, for example by explicitly mentioning authors or sources (e.g., ``Author et al. argue that ...'') to improve contextual clarity~\cite{shaier-etal-2024-adaptive}. A different approach is the \textbf{highlight gradient}, which visually encodes the influence of cited content by coloring relevant tokens or sentences in both the generated output and the source text, thereby providing a visual form of traceability~\cite{Do2024facilitatinghumanllmcollaborationfactuality}. Finally, we identify a single paper that employs direct \textbf{quotes}, embedding verbatim excerpts from the evidence source into the generated response~\cite{xiao2025quillquotationgenerationenhancement}. Importantly, quotations are not limited to serving as a citation style. In our analysis, we find that 13\% of the reviewed studies explicitly engage with quotations in some form. Moreover, quotations can function as a training or prompting strategy, as exemplified by approaches such as chain-of-quote \cite{li-etal-2024-making}. In their user study, \citet{Do2024facilitatinghumanllmcollaborationfactuality} show that \ac{llm} responses with citations are perceived as significantly more trustworthy than uncited responses, while no significant difference in trust is observed between in-line citations and highlight gradient visualizations.

\textbf{Task-specific Analysis.} In-line citations are among the top-2 most frequently used citation style across all tasks, only for grounded text generation and fact verification citation reports appear slightly more often, making them the default style for presenting evidence in \ac{llm}-generated text. \ac{qa} exhibits the greatest diversity of citation styles, including in-line citations, citation reports, passages, narrative citations, and highlight gradients. This diversity reflects the overall frequency of studies for \ac{qa}, but also the wide range of different settings of this task. Grounded text generation shows a similar diversity of citation styles, but being currently the only task where quotes have been utilized. Summarization, citation text generation, and related work generation rely primarily on in-line citations, largely following conventions from document-centric and scientific writing, where citations are embedded directly in the text. While fact verification studies predominantly use citation reports, the limited sample size prevents drawing reliable conclusions about preferred citation mechanisms in this task. Overall, citation style choices show only limited task-specific variation. Across tasks, most studies employ fine-grained citation styles, such as in-line citations, that allow users to verify individual \ac{llm}-generated claims, while differences related to task requirements or user interaction remain marginal.

\textbf{Trends.} Citation styles exhibit a staggered emergence over time, with clear differences in adoption. Passages appear earliest, first emerging in Q2 2021, but grow only moderately. In-line citations are introduced later, first appearing in Q1 2022, and show the strongest growth trajectory, increasing substantially from 11 studies in 2023 to 67 in 2024. Citation reports also emerge in late 2022 and expand steadily, rising from 10 publications in 2023 to 19 in 2024. Narrative citations appear only from Q3 2023 onward and remain comparatively rare, while highlight gradients and quotes emerge only in 2024 and are explored by very few studies. Overall, the literature shows a rapid consolidation around a small set of dominant citation styles, with more specialized citation styles appearing only recently and at limited scale.

\subsubsection{Citation Visibility}
\label{sec:additional_citation_visibility}

Citation visibility describes whether citations are visible to users in the \ac{llm}-generated output, or are generated only internally in intermediate text. The majority of reviewed papers include citations directly in the \textbf{final response}, making supporting evidence visible to users and enabling explicit traceability for verification. Overall, 91\% of the surveyed studies adopt this form of citation visibility. In contrast, 4\% of studies generate citations exclusively in \textbf{intermediate text} that is not shown to users. In these cases, citations are used internally by the \ac{llm} as part of the generation or evaluation process. Across the reviewed studies, citations in intermediate text serve three main purposes: \textit{internal verification}~\cite{fang-etal-2024-hgot, chiang-lee-2024-merging}, \textit{guiding multi-step reasoning during generation}~\cite{chu-etal-2025-towards, Xia2025RALMSelfReasoning}, and \textit{aligning evidence across modalities in multimodal settings}~\cite{suri-etal-2025-visdom}.

\textbf{Task-specific Analysis.} Citation visibility differs across tasks but is dominated by citations in the final response. Summarization, citation text generation, related work generation, and fact verification exclusively rely on citations that are visible to users. This aligns with the objective of these tasks, which is to explicitly present evidence to support generated content. \ac{qa} and grounded text generation are the only tasks that also include approaches using citations in intermediate text. In these cases, citations are used internally to optimize aspects such as answer correctness, without necessarily exposing supporting evidence to users.

\textbf{Trends.} Citations shown in the final response have been present since the earliest work on evidence-based text generation, with the first studies appearing in Q2 2021. The number of approaches exposing citations to users increased substantially from 24 papers in 2023 to 79 in 2024. In contrast, citation visibility limited to intermediate text emerged only recently, first appearing in Q4 2023, and remains rare, with only a small number of studies in 2023 and 2024. Overall, trends indicate strong growth in citations in the final response, while intermediate text citations remain a sparsely utilized design choice.

\subsubsection{Citation Frequency}
\label{sec:additional_citation_frequency}

Citation frequency describes the number of citations provided for each \ac{llm}-generated claim. We distinguish between approaches that provide a \textbf{single} citation per claim and those that support \textbf{multiple} citations. Overall, 64\% of the reviewed studies support multiple citations per claim, indicating a preference for broader evidential coverage. This design choice reflects a trade-off between evidential coverage and simplicity. Multiple citations can strengthen support by drawing on diverse sources, whereas single citations (31\% of studies) offer a more concise presentation and may reduce cognitive load for users. We observe that some parametric approaches are limited to generating a single citation per claim due to architectural constraints~\cite{Khalifa2024SourceAwareTraining, lu-etal-2025-wasa}. In contrast, most non-parametric approaches naturally support multiple citations by aggregating evidence through retrieval mechanisms. \citet{Ding2025CitationsTrust} found that while the presence of citations increases perceived trust in \ac{llm}-generated responses, adding more than one citation does not yield additional trust gains.

\textbf{Task-specific Analysis.} While all tasks include approaches that support multiple citations per \ac{llm}-generated claim, \ac{qa}, grounded text generation, summarization, and fact verification predominantly rely on multiple citations, reflecting the need to aggregate evidence from several sources. In contrast, citation text generation and related work generation more often rely on a single citation per claim. This reflects differences in task formulation. In citation text generation, a single paper is typically provided to the \ac{llm} and incorporated into an existing text, naturally resulting in a single citation per claim. In related work generation, the goal is to discuss and compare individual papers, with claims commonly tied to specific works rather than to an aggregation of sources.

\textbf{Trends.} Studies supporting only single citations per claim appear earlier, with the first paper published in Q2 2021. Support for multiple citations emerges in Q4 2022, but shows substantially stronger growth. The number of approaches supporting multiple citations increased from 18 in 2023 to 64 in 2024, compared to an increase from 6 to 29 for single-citation approaches over the same period. Overall, while single citations remain preferred in certain task formulations, the ability to provide multiple citations has become widely supported and no longer represents a limiting factor for evidence-based text generation with \acp{llm}.

\subsection{Task}
\label{sec:additional_task}

In this section, we summarize the distribution of tasks studied in evidence-based text generation with \acp{llm}. Across the literature, we identified 16 distinct tasks, of which only six are addressed by more than one study. As shown in Table~\ref{tab:task-overview-table}, \ac{qa} dominates the field, while grounded text generation and summarization receive moderate attention. Other tasks, including citation text generation, related work generation, and fact verification, are investigated less frequently.

\begin{table}[t]
    \centering
    \small
    \begin{tabular}{l c}
    \toprule
    \textbf{Task} & \textbf{No. of Papers} \\
    \midrule
    Question Answering & 95 \\
    Grounded Text Generation & 15 \\
    Summarization & 14 \\
    Citation Text Generation & 6 \\
    Related Work Generation & 5 \\
    Fact Verification & 3 \\
    \bottomrule
    \end{tabular}
    \caption{Overview of the six most popular tasks in the literature on evidence-based text generation with \acp{llm}. Each paper can be assigned to multiple tasks, so totals exceed 134.}
    \label{tab:task-overview-table}
\end{table}

\textbf{Trends.} Evidence-based text generation tasks exhibit distinct temporal adoption patterns. The earliest task in our annotated dataset is grounded text generation, which first appears in Q2 2021. After limited activity for several years, it gains momentum only recently, increasing from 2 publications in 2023 to 11 in 2024. \ac{qa} emerges next, first appearing in Q1 2022 and subsequently becoming the dominant task. The number of papers in this category rise sharply from 18 in 2023 to 71 in 2024, reflecting its central role in evidence-based text generation with \acp{llm}. Summarization enters the landscape in Q4 2022 and shows notable growth, increasing from 2 studies in 2023 to 11 in 2024. Citation text generation appears somewhat later, with three studies in 2023 and two in 2024, indicating steady but comparatively modest activity. Related work generation and fact verification are the most recent tasks, both first emerging in Q4 2023. These areas show early signs of uptake, with related work generation growing from one study in 2023 to four in 2024, and fact verification from one to two. Overall, the task-level trends reveal a broadening scope of evidence-based text generation with \acp{llm}, with \ac{qa} driving much of the recent growth and grounded text generation and summarization also showing increased adoption.

\section{Evaluation Resources}
\label{sec:evaluation_resources}

This section provides additional details on the evaluation of evidence-based text generation with \acp{llm}, complementing Section~\ref{sec:evaluation}. We define evaluation frameworks as standardized procedures that systematically apply a predefined set of metrics (at least two) over specific tasks to assess methodological performance along at least one evaluation dimension. In contrast, benchmarks are curated combinations of datasets that facilitate reproducible and comparative evaluation, while individual metrics and datasets are not assigned to a framework or benchmark. Overall, we identify 19 evaluation frameworks and 11 benchmarks, of which only two from each group are reused across multiple studies, highlighting a persistent lack of consensus and standardization in current evaluation practices.

Section~\ref{sec:appendix_evaluation_methods} provides task-specific analyses and identifies trends across evaluation dimensions, while Section~\ref{sec:appendix_evaluation_dimensions} presents a complementary perspective of evaluation dimensions. In addition, Section~\ref{sec:additional_metrics} discusses notable evaluation metrics that do not appear among the frequently reused metrics detailed in Section~\ref{sec:evaluation_dimensions}. Finally, Section~\ref{sec:frameworks} and Section~\ref{sec:benchmarks} summarize the identified frameworks and benchmarks, respectively, and Section~\ref{sec:datasets} provides an overview of available evaluation datasets.

\subsection{Evaluation Methods}
\label{sec:appendix_evaluation_methods}

We identified six evaluation methods, described in Section~\ref{sec:evaluation_methods}. Table~\ref{tab:evaluation_method} summarizes these evaluation methods along with supplementary details not included in the main text.

\begin{figure*}[ht!]
    \centering
    \includegraphics[width=\linewidth]{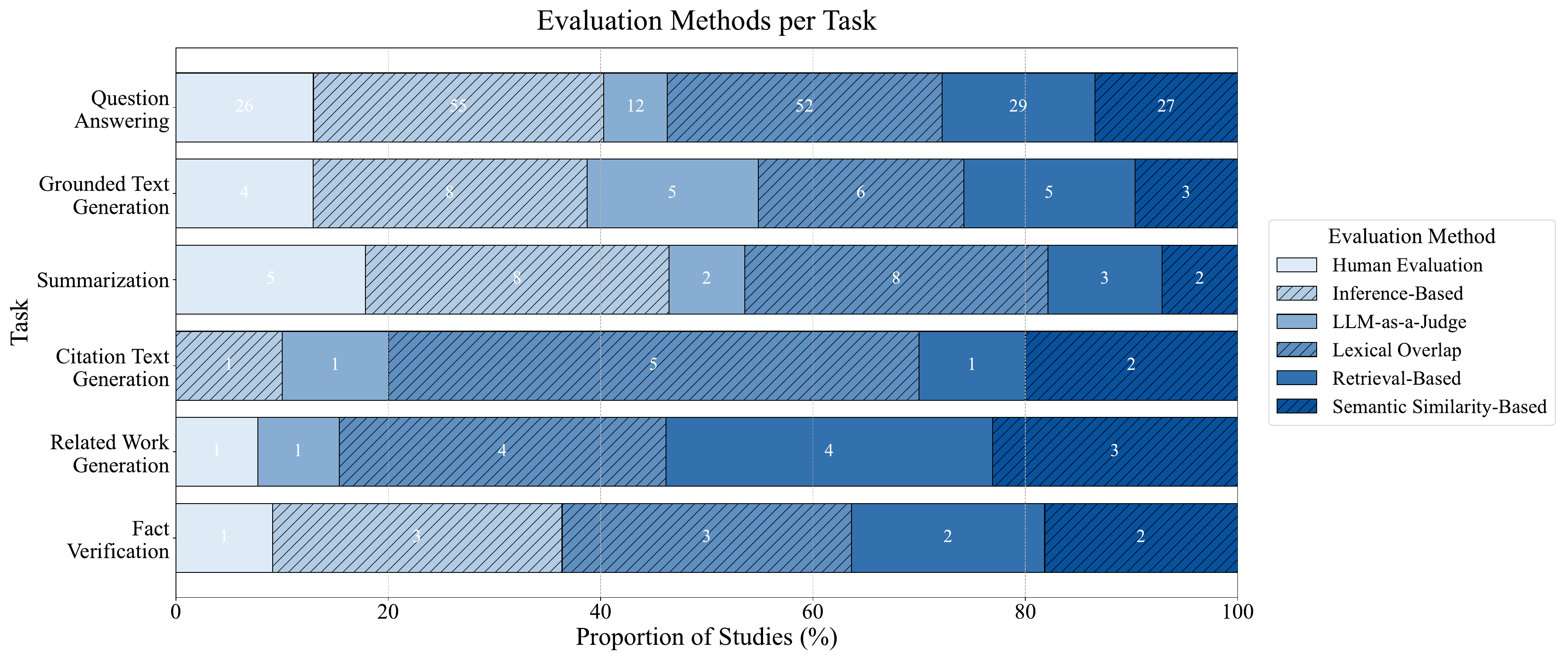}
    \caption{Distribution of evaluation methods across tasks in evidence-based text generation with \acp{llm}. The bars represent the relative percentages of different evaluation methods for each task. The absolute number of papers for each evaluation method is reported alongside the bars. Tasks include question answering, grounded text generation, summarization, citation text generation, related work generation, and fact verification. Each paper can be assigned to multiple tasks, so totals exceed 134.}
    \label{fig:evaluation_methods_per_task}
\end{figure*}

\textbf{Task-specific Analysis.} Figure \ref{fig:evaluation_methods_per_task} shows how evaluation methods distribute across tasks. \ac{qa}, grounded text generation, and summarization exhibit highly similar evaluation patterns. Across these tasks, evaluation is dominated by inference-based and lexical overlap methods, which together account for 53\% of evaluation usage in \ac{qa}, 45\% in grounded text generation, and 57\% in summarization. This indicates a strong reliance on automatic metrics, while human evaluation represents the largest relative share of evaluation in summarization, highlighting the limitations of purely automatic methods. In contrast, \ac{llm}-as-a-judge evaluation remains relatively uncommon across tasks, reaching its highest share in grounded text generation (16\%) while remaining below 10\% in all other tasks, indicating that \ac{llm}-as-a-judge evaluation is still emerging and has not yet replaced established automatic approaches.
Citation text generation and related work generation show a distinct evaluation pattern. In citation text generation, lexical overlap methods clearly dominate, accounting for 50\% of evaluation usage. In related work generation, lexical overlap and retrieval-based methods are most prevalent, together accounting for 62\% of evaluation usage.
Fact verification exhibits the narrowest distribution of evaluation methods, with inference-based and lexical overlap approaches accounting for 55\% of evaluation usage, while the absence of \ac{llm}-as-a-judge evaluation indicates continued reliance on deterministic and interpretable methods. 
Overall, the observed patterns indicate that traditional automated metrics, particularly lexical overlap and semantic similarity-based methods, continue to dominate the evaluation of shorter text segments, while long-form text is assessed mainly using inference-based methods, with \ac{llm}-as-a-judge approaches still used only to a limited extent.

\textbf{Trends.} Evaluation methods exhibit distinct temporal adoption patterns. The earliest methods in our annotated dataset are inference-based, lexical overlap, and semantic similarity-based metrics, all first appearing in Q2 2021. These traditional automated metrics have since experienced substantial growth. Inference-based evaluation shows the largest absolute increase, rising from 11 studies in 2023 to 51 in 2024, while lexical overlap grows from 14 to 50 over the same period. Semantic similarity-based evaluation also expands strongly, increasing from 5 to 29 studies during this interval. 
Human evaluation was applied slightly later, first appearing in Q1 2022, and shows steady growth, increasing from 8 to 23 studies over the same period. More recent evaluation paradigms appear later in the timeline. \ac{llm}-as-a-judge evaluation first emerges in Q4 2022 and grows moderately from 5 to 17 studies. Retrieval-based evaluation is the most recent approach in this domain, first appearing in Q3 2023, although the underlying methodology is well established. Nevertheless, it exhibits the fastest growth overall, increasing from 4 to 33 studies during this interval. Overall, these temporal patterns indicate that evaluation practices remain strongly anchored in traditional automated metrics, while newer paradigms that leverage retrieval signals or \acp{llm} for evaluation are emerging but still less widely adopted.

\subsection{Evaluation Dimensions}
\label{sec:appendix_evaluation_dimensions}

\textbf{Task-specific Analysis.} Evaluation dimensions show a high degree of consistency across tasks. The three core dimensions (attribution, citation, and correctness) clearly dominate evaluation practices in all tasks, together accounting for between 64\% and 81\% of evaluation instances. The evaluation of attribution and correctness is prominent across all tasks, while the evaluation dimension citation is applied whenever annotated ground-truth evidence is available. Notably, citation has not yet been evaluated in fact verification, where annotated ground-truth evidence is often unavailable, and studies therefore rely primarily on attribution and correctness.
In contrast, contextual evaluation dimensions exhibit greater variation across tasks. Linguistic quality is the most frequently applied contextual dimension and appears across most tasks, whereas preservation, relevance, and retrieval are used only selectively. These dimensions reflect task-specific evaluation needs rather than core requirements for verifying evidence-based text generation with \acp{llm}.
Overall, the observed patterns indicate strong adherence to the core evaluation dimensions defined in our guidelines. Attribution, citation, and correctness consistently form the foundation of evaluation across tasks, while contextual dimensions serve a complementary and task-dependent role.

\textbf{Trends.} Evaluation dimensions have been applied at different stages over time. The earliest dimension in our annotated dataset is attribution, first appearing in Q2 2021, followed by correctness in Q1 2022. These core evaluation dimensions have since experienced substantial growth. Correctness shows the largest absolute increase, rising from 13 studies in 2023 to 63 in 2024, while attribution grows from 16 to 60 over the same period. The evaluation of citation is applied later, first appearing in Q4 2022, but increases from 5 to 30 studies during this interval.
Contextual evaluation dimensions are applied at later stages. Linguistic quality, first introduced in Q4 2022, shows steady growth from 7 to 31 studies over the same period. Preservation remains rarely applied, increasing only marginally from 1 to 3 studies. Relevance and retrieval are the most recent dimensions, both first appearing in Q3 2023, and show moderate growth from 2 to 13 studies.
Overall, these temporal patterns indicate that evaluation practices remain strongly anchored in core dimensions, particularly attribution and correctness, while contextual dimensions have emerged more recently and continue to be applied more selectively depending on task requirements.

\subsection{Additional Notable Evaluation Metrics}
\label{sec:additional_metrics}

Given the total of 300 extracted metrics, a detailed description of each metric is beyond the scope of this survey. To ensure clarity and encourage evaluation standardization, Section~\ref{sec:evaluation_dimensions} highlights the most frequently reused metrics. We acknowledge that some relevant, especially recent, evaluation metrics may not yet meet this criterion. Although not yet widely adopted, these additional evaluation metrics are still significant for the evolving evaluation landscape. Accordingly, this section presents further notable evaluation metrics that may shape future research. The full table of 300 categorized evaluation metrics is available in our repository, as described in Appendix~\ref{sec:availability}.

\textbf{Attribution.} Early work in multilingual \ac{qa} attribution by \citet{muller-etal-2023-evaluating} introduced \textit{Attr-Acc}, an inference-based metric to evaluate the classification accuracy of attribution detection models. Building on this, \citet{yue-etal-2023-automatic} proposed \textit{AttrScore}, which extends binary attribution accuracy to a three-way classification, distinguishing whether a generated statement is fully supported, insufficiently supported, or contradicted by a cited reference. In the context of citation text generation, \citet{sahinuc-etal-2024-systematic} revealed relationships between different metrics, including \textit{ROUGE-L}, \textit{BERT-Score}, \textit{SciBERT-Score}, \textit{BLEURT}, and \textit{SummaC}. Extending attribution evaluation to a finer evidence level, \citet{phukan-etal-2024-peering} proposes the metrics \textit{Token-level Recall}, \textit{Precision}, and \textit{F1}, which measure how accurately and completely the model identifies the correct source tokens, as well as \textit{Span-level Accuracy}, which assesses the percentage of predicted spans that exactly match the ground truth. More recent work by \citet{yan2024atomicfactdecompositionhelps} proposes decomposing claims in \ac{llm}-generated answers into atomic facts and defines the inference-based metrics \textit{Attribution Recall} and \textit{Precision} to quantify the proportion of claims supported by evidence and the relevance of the evidence retrieved, respectively. Leveraging knowledge graphs to generate comprehensive attributions, \textit{CAQA} enables \citet{hu-etal-2025-llms} to evaluate 25 automatic and human evaluators, as well as \acp{llm} fine-tuned on the task. Separately, \citet{xing-etal-2025-evaluating} introduce the metrics \textit{Set Precision}, \textit{Recall}, and \textit{F1}, which assess how accurately an annotator, human or \ac{llm}, identifies citations within a masked explanation. For long-form Wikipedia generation, \citet{zhang-etal-2025-wikigenbench} introduce the inference-based metric \textit{Citation Rate}, defined as the word-length-weighted average of sentence-level Citation Recall \ac{nli}, which accounts for variation in sentence length. \citet{Qian2025CitationGeneration} additionally extend the ALCE framework by using GPT models as \ac{nli} component, observing similar overall trends to classical \ac{nli} models but generally lower citation scores with the GPT-based approach. \citet{Wallat2025CorrectnessFaithfulness} report that up to 57\% of citations generated by \ac{rag} systems are unfaithful, and suggest that attribution evaluation should also consider faithfulness, that is, whether sources were actually used during text generation, as attribution metrics primarily capture whether sources support a generated claim.

\textbf{Correctness.} Beyond the metrics presented in Section~\ref{sec:evaluation_dimensions}, several additional studies have proposed notable approaches. \citet{chiang-lee-2024-merging} propose \textit{D-FActScore}, an extension of \textit{FActScore} \cite{min-etal-2023-factscore} designed to better handle entity ambiguity. Whereas FActScore evaluates each fact independently, D-FActScore groups related facts that a reader could reasonably interpret as referring to the same entity without prior knowledge.

\textbf{Citation.} To evaluate citations, \citet{nishimura-etal-2024-toward} complement existing metrics with \textit{ARI'}, a modification of the Adjusted Rand Index that accounts for unmatched citations across multiple paragraphs, and \textit{Citation Structure F1}, which measures how closely the citation patterns in generated paragraphs match those in the ground truth. In long-context summarization, \citet{laban-etal-2024-summary} proposed a \textit{Citation Score}, measuring citation accuracy, and a \textit{Coverage Score}, using \ac{llm}-as-a-judge to assess correctness. They also introduced a \textit{Joint Score}, combining both metrics to evaluate whether \acp{llm} and \ac{rag} systems preserve comprehensive coverage of cited content.

\textbf{Preservation.} \citet{sahinuc-etal-2024-systematic} employed \textit{N-Gram Overlap}, a lexical overlap metric, to measure n-gram overlap between the input and the model output in citation text generation, assessing the extent of lexical reuse from the prompt.

\subsection{Frameworks} 
\label{sec:frameworks}

Figure~\ref{fig:evaluation} maps the two evaluation frameworks that have been reused for evidence-based text generation with \acp{llm} to the dimensions they address. The first, \textit{ALCE}~\cite{gao-etal-2023-enabling}, is the most widely adopted, appearing in 12 studies and covering attribution, correctness, and linguistic quality. Beyond applying the entire framework, several additional studies use only individual ALCE metrics, such as Citation Precision \ac{nli} and Citation Recall \ac{nli}, without adopting the complete framework. While these partial uses are not counted among the 12 instances adopting the framework, they further underscore the influence and prominence of ALCE in current evaluation practices. The second framework, \textit{G-Eval}~\cite{liu-etal-2023-g}, appears in two studies and provides a broader evaluation of natural language generation through an \ac{llm}-as-a-judge approach, assessing attribution, linguistic quality, and relevance. Although only ALCE and G-Eval have seen reuse so far, several additional frameworks have been introduced since 2023. While not yet adopted in published research, they remain relevant to the evaluation landscape. Table~\ref{tab:frameworks} therefore presents a complete overview of all frameworks, both reused and not yet adopted, detailing their respective evaluation dimensions and methods.

In addition to these evaluation frameworks, \citet{zhang-etal-2024-towards-fine-grained} propose a meta-framework that systematically evaluates the effectiveness of citation and faithfulness metrics themselves, highlighting that not only evaluation frameworks but also meta-frameworks have emerged in this area.

\subsection{Benchmarks} 
\label{sec:benchmarks}

Similar to the evaluation frameworks, we identified only two benchmarks that were used more than once to evaluate evidence-based text generation approaches. \textit{ALCE}~\cite{gao-etal-2023-enabling} functions both as a framework and benchmark, and has been reused as a benchmark in nine studies. This benchmark supports multiple evaluation dimensions and is built on datasets focused on information-seeking tasks. \textit{RAG-RewardBench}~\cite{jin-etal-2025-rag}, designed to evaluate reward models for preference alignment in \ac{rag}, is used twice. It spans 18 datasets and targets evaluation scenarios such as multi-hop reasoning, fine-grained citation, appropriate abstention, and conflict robustness. Although only two benchmarks have been reused, several others have been introduced in the literature. While these have not yet been reused in published research, they remain relevant within the evaluation landscape. Therefore, Table~\ref{tab:benchmarks} provides the full list of evaluation benchmarks, including their associated task and data domain.

\subsection{Datasets} 
\label{sec:datasets}

Overall, we extracted 231 highly task-specific datasets serving diverse purposes, including both training and evaluation. Due to this heterogeneity, we focus on the most frequently reused datasets that support standardized evaluation, so far primarily centered on \ac{qa} with more than 64\% of datasets, followed by grounded text generation (9\%) and summarization (6\%). Table~\ref{tab:datasets} presents the 12 most frequent datasets for \ac{qa}, the eight most frequent datasets for grounded text generation, and the four most frequent datasets for all other tasks used to evaluate evidence-based text generation with \acp{llm}. Additionally, we identified the data domain for each frequent dataset and found that most datasets fall within the Wikipedia, scientific, and news domains, while others are less frequently considered. The full list of datasets is available in our Git repository, as described in Appendix~\ref{sec:availability}.

\begin{table*}[t]
\centering
\scriptsize
\setlength{\tabcolsep}{4pt}
\renewcommand{\arraystretch}{1.2}

\begin{tabular}{|p{3.5cm}|p{3.5cm}|p{3.5cm}|p{3.5cm}|}
\hline

\textbf{Contribution Type} & \textbf{Parametric} & \textbf{Non-Parametric} & \textbf{Citation Modality} \\
\hline
APP: Approach & PLL: Pure \ac{llm} & PR: Post-Retrieval & TX: Texts \\
EVA: Evaluation & MC: Model-Centric & PG: Post-Generation & GR: Graphs \\
RES: Resource & DC: Data-Centric & IG: In-Generation & TB: Tables \\
APL: Application &  & IC: In-Context & VL: Visuals \\
POS: Position &  &  &  \\
SUR: Survey &  &  &  \\
\hline

\textbf{Evidence Level} & \textbf{Citation Style} & \textbf{Citation Visibility} & \textbf{Citation Frequency} \\
\hline
DOC: Document & ILC: In-Line Citations & FR: Final Response & SI: Single \\
PAR: Paragraph & CR: Citation Report & IT: Intermediate Text & MU: Multiple \\
SEN: Sentence & PAS: Passage &  &  \\
TOK: Token & NC: Narrative Citations &  &  \\
TRI: Triple & HG: Highlight Gradient &  &  \\
TAB: Table & QU: Quote &  &  \\
TC: Table Cell &  &  &  \\
IMG: Image &  &  &  \\
BB: Bounding Box &  &  &  \\
\hline

\textbf{Task} & \textbf{Training} & \textbf{Prompting} & \textbf{Evaluation Dimension} \\
\hline
QA: Question Answering & SFT: Supervised Fine-Tuning & ZS: Zero-Shot & ATT: Attribution \\
GTG: Grounded Text Generation & SSFT: Self-Supervised Fine-Tuning & FS: Few-Shot & COR: Correctness \\
SUM: Summarization & RL: Reinforcement Learning & COT: Chain-Of-Thought & CIT: Citation \\
RWG: Related Work Generation & PT: Pretraining & SC: Self-Consistency & LQ: Linguistic Quality \\
CTG: Citation Text Generation &  & AO: Active-Oriented & PRE: Preservation \\
FV: Fact Verification &  & COC: Chain-Of-Citation & REL: Relevance \\
OTH: Other &  & COQ: Chain-Of-Quote & RET: Retrieval \\
 &  & CA: Conflict-Aware &  \\
 &  & RP: Role-Play &  \\
\hline

\textbf{Evaluation Method} & & & \\
\hline
HE: Human Evaluation & & & \\
IB: Inference-Based & & & \\
LO: Lexical Overlap & & & \\
LJ: \ac{llm}-as-a-Judge & & & \\
RB: Retrieval-Based & & & \\
SB: Semantic Similarity-Based & & & \\
\hline

\end{tabular}

\caption{Legend of all abbreviations used for contribution types, attribution approaches, citation characteristics, task categories,  \ac{llm} integration strategies, evaluation dimensions, and evaluation methods. These abbreviations are used throughout Table~\ref{tab:classification_papers}, Table~\ref{tab:metric-comparison}, Table~\ref{tab:frameworks}, Table~\ref{tab:benchmarks}, and Table~\ref{tab:datasets}.}
\label{tab:legend}
\end{table*}

\clearpage
\onecolumn
\begingroup
\begin{scriptsize}
\captionsetup{skip=0pt}
\begin{longtable}{l|l|ll|lllll|l|ll}
\toprule
\textbf{Paper} & \makecell[l]{\textbf{Contribution} \\ \textbf{Type}}
& \multicolumn{2}{l|}{\makecell[l]{\textbf{Attribution} \\ \textbf{Approach}}}
& \multicolumn{5}{l|}{\makecell[l]{\textbf{Citation} \\ \textbf{Characteristics}}}
& \textbf{Task}
& \multicolumn{2}{l}{\makecell[l]{\textbf{\ac{llm}} \\ \textbf{Integration}}} \\
\cmidrule(lr){3-4} \cmidrule(lr){5-9} \cmidrule(lr){11-12}
& & Par. & Non-Par. & Mod. & Evi. & Style & Vis. & Freq. & & Train. & Prompt. \\
\midrule
\endfirsthead

\toprule
\textbf{Paper} & \makecell[l]{\textbf{Contribution} \\ \textbf{Type}}
& \multicolumn{2}{l|}{\makecell[l]{\textbf{Attribution} \\ \textbf{Approach}}}
& \multicolumn{5}{l|}{\makecell[l]{\textbf{Citation} \\ \textbf{Characteristics}}}
& \textbf{Task}
& \multicolumn{2}{l}{\makecell[l]{\textbf{\ac{llm}} \\ \textbf{Integration}}} \\
\cmidrule(lr){3-4} \cmidrule(lr){5-9} \cmidrule(lr){11-12}
& & Par. & Non-Par. & Mod. & Evi. & Style & Vis. & Freq. & & Train. & Prompt. \\
\midrule
\endhead

\midrule
\endfoot

\citet{dziri-etal-2022-evaluating} & EVA, RES & – & – & TX & PAR & PAS & FR & SI & GTG & SFT & – \\
\citet{menick2022teachinglanguagemodelssupport} & APP & – & PR & TX & \makecell[l]{DOC,\\ SEN} & ILC & FR & SI & QA & \makecell[l]{SFT,\\ RL} & FS \\
\citet{gao-etal-2023-rarr} & APP & – & PG & TX & PAR & CR & FR & MU & \makecell[l]{QA,\\ SUM} & SFT & FS \\
\citet{gu-hahnloser-2024-controllable} & APP, RES & – & IC & TX & DOC & ILC & FR & SI & CTG & \makecell[l]{SFT,\\ RL} & ZS \\
\citet{bohnet2023attributedquestionansweringevaluation} & EVA, RES & – & \makecell[l]{PR,\\ PG} & TX & DOC & CR & FR & SI & QA & SFT & FS \\
\citet{yue-etal-2023-automatic} & EVA & – & – & TX & DOC & ILC & FR & SI & QA & SFT & \makecell[l]{ZS,\\ FS} \\
\citet{gao-etal-2023-enabling} & \makecell[l]{APP, EVA,\\ RES} & – & \makecell[l]{PR,\\ PG} & TX & PAR & ILC & FR & MU & QA & – & ZS \\
\citet{muller-etal-2023-evaluating} & \makecell[l]{APP, EVA,\\ RES} & – & PR & TX & PAR & PAS & FR & SI & QA & SFT & \makecell[l]{FS,\\ COT} \\
\citet{chen2023purrefficientlyeditinglanguage} & APP, RES & – & PR & TX & PAR & CR & FR & MU & GTG & SSFT & \makecell[l]{ZS,\\ FS,\\ COT} \\
\citet{Pride2023CORE-GPT} & EVA, APL & PLL & PR & TX & DOC & CR & FR & MU & QA & – & ZS \\
\citet{huang-chang-2024-citation} & POS & \makecell[l]{MC,\\ DC} & \makecell[l]{PR,\\ PG} & TX & – & ILC & FR & – & CTG & – & – \\
\citet{kamalloo2023hagridhumanllmcollaborativedataset} & RES & – & – & TX & – & – & – & MU & QA & – & – \\
\citet{Kang2023Synergi} & APL & – & PR & TX & DOC & \makecell[l]{ILC,\\ CR,\\ NC} & FR & MU & SUM & – & ZS \\
\citet{xue2024weaverbirdempoweringfinancialdecisionmaking} & APL & – & PR & TX & DOC & CR & FR & MU & QA & SFT & ZS \\
\citet{Spennemann2025Veracity} & EVA & PLL & – & TX & DOC & CR & FR & MU & QA & – & – \\
\citet{Zuccon2023ChatGPTCitation} & EVA & PLL & – & TX & DOC & CR & FR & MU & QA & – & ZS \\
\citet{Hu2025GeneSetFunctions} & EVA, APL & PLL & – & TX & DOC & CR & FR & MU & OTH & – & ZS \\
\citet{malaviya-etal-2024-expertqa} & \makecell[l]{APP, EVA,\\ RES} & PLL & \makecell[l]{PR,\\ PG} & TX & DOC & ILC & FR & MU & QA & SFT & ZS \\
\citet{lee2023reliablefluentlargelanguage} & APP, RES & – & PR & TX & PAR & ILC & FR & MU & QA & SFT & ZS \\
\citet{zhang2023largelanguagemodelsmeet} & SUR & – & – & TX & – & – & – & – & OTH & – & – \\
\citet{Fok2024Qlarify} & APL & – & PR & TX & \makecell[l]{DOC,\\ PAR} & PAS & FR & SI & \makecell[l]{QA,\\ SUM} & – & \makecell[l]{ZS,\\ FS} \\
\citet{asai2024selfrag} & APP & – & IG & TX & PAR & CR & FR & MU & \makecell[l]{QA,\\ FV} & SFT & ZS \\
\citet{lu-etal-2025-wasa} & APP & DC & – & TX & SEN & PAS & FR & SI & GTG & \makecell[l]{SFT,\\ PT} & – \\
\citet{li-etal-2024-towards-verifiable} & EVA, RES & – & PR & GR & TRI & ILC & FR & MU & QA & – & FS \\
\citet{li2023surveylargelanguagemodels} & SUR & – & – & TX & – & – & – & – & OTH & – & – \\
\citet{ye-etal-2024-effective} & APP & – & IG & TX & PAR & ILC & FR & MU & QA & SFT & ZS \\
\citet{schuster-etal-2024-semqa} & APP, RES & – & PR & TX & DOC & ILC & FR & MU & QA & SFT & \makecell[l]{ZS,\\ FS} \\
\citet{Worledge2024Unifying} & APP & – & – & – & – & – & FR & – & QA & – & – \\
\citet{malaviya-etal-2024-said} & EVA & – & IC & TX & SEN & PAS & IT & SI & QA & – & FS \\
\citet{Anand2023CitationTextGeneration} & APP, RES & – & IC & TX & DOC & CR & FR & MU & CTG & SFT & ZS \\
\citet{Anand2023KG-CTG} & APP & – & IC & TX & DOC & CR & FR & SI & CTG & SFT & ZS \\
\citet{Agarwal2025LitLLMs} & APP, RES & – & PR & TX & DOC & ILC & FR & MU & RWG & SFT & ZS \\
\citet{sun-etal-2024-towards-verifiable} & APP & – & PR & TX & PAR & ILC & FR & MU & QA & – & ZS \\
\citet{hu-etal-2025-llms} & EVA, RES & – & PR & TX & DOC & ILC & FR & MU & QA & SFT & ZS \\
\citet{li-etal-2024-attributionbench} & EVA, RES & – & – & TX & PAR & – & – & SI & OTH & – & – \\
\citet{li-etal-2024-citation} & APP & – & PG & TX & PAR & ILC & FR & SI & QA & – & \makecell[l]{ZS,\\ COT} \\
\citet{li-ouyang-2025-explaining} & APP & – & \makecell[l]{PR,\\ IC} & TX & DOC & NC & FR & SI & RWG & – & ZS \\
\citet{fang-etal-2024-hgot} & APP, EVA & – & PR & TX & PAR & ILC & IT & MU & QA & – & FS \\
\citet{chiang-lee-2024-merging} & APP, EVA & – & PR & TX & PAR & ILC & IT & MU & GTG & – & FS \\
\citet{huang-etal-2024-training} & APP & – & PR & TX & PAR & ILC & FR & MU & QA & \makecell[l]{SFT,\\ RL} & ZS \\
\citet{zhang-etal-2025-wikigenbench} & EVA, RES & – & PR & TX & DOC & ILC & FR & MU & SUM & SFT & ZS \\
\citet{lee-etal-2024-ask} & APP, EVA & – & IC & TX & PAR & ILC & FR & MU & QA & – & ZS \\
\citet{slobodkin-etal-2024-attribute} & APP & – & IC & TX & SEN & ILC & FR & MU & \makecell[l]{QA,\\ SUM} & SFT & \makecell[l]{FS,\\ COT} \\
\citet{li-etal-2024-improving-attributed} & APP, RES & DC & – & TX & PAR & ILC & FR & MU & QA & \makecell[l]{SFT,\\ RL} & ZS \\
\citet{deng-etal-2024-webcites} & \makecell[l]{APP, EVA,\\ RES} & – & PR & TX & PAR & ILC & FR & MU & \makecell[l]{QA,\\ SUM} & SFT & \makecell[l]{ZS,\\ FS} \\
\citet{berchansky-etal-2024-cotar} & APP & – & PR & TX & PAR & ILC & FR & MU & QA & SFT & FS \\
\citet{Li2025SurveyGenerativeIR} & SUR & – & – & TX & – & – & – & – & OTH & – & – \\
\citet{Lee2025LLMAttributor} & APL & PLL & – & TX & \makecell[l]{DOC,\\ PAR,\\ TOK} & \makecell[l]{CR,\\ HG} & FR & MU & QA & SFT & – \\
\citet{fierro-etal-2024-learning} & APP & – & PR & TX & PAR & ILC & FR & MU & \makecell[l]{QA,\\ SUM} & SFT & – \\
\citet{Khalifa2024SourceAwareTraining} & APP & MC & – & TX & DOC & ILC & FR & SI & QA & \makecell[l]{SFT,\\ PT} & – \\
\citet{golany-etal-2024-efficient} & RES, APL & – & IC & TX & PAR & CR & FR & MU & GTG & SFT & FS \\
\citet{Do2024facilitatinghumanllmcollaborationfactuality} & EVA & PLL & – & TX & PAR & \makecell[l]{ILC,\\ PAS,\\ HG} & FR & MU & OTH & – & – \\
\citet{magesh2024hallucinationfreeassessingreliabilityleading} & EVA, RES & PLL & PR & TX & DOC & ILC & FR & MU & QA & – & – \\
\citet{li2024nearest} & APP & – & \makecell[l]{PR,\\ IG} & TX & PAR & ILC & FR & SI & \makecell[l]{QA,\\ FV,\\ OTH} & – & ZS \\
\citet{Mayfield2024Reports} & POS & – & PR & TX & DOC & ILC & FR & MU & SUM & – & – \\
\citet{phukan-etal-2024-peering} & \makecell[l]{APP, EVA,\\ RES} & – & – & TX & TOK & PAS & FR & SI & QA & – & – \\
\citet{maheshwari-etal-2024-presentations} & APP, EVA & – & IC & TX & PAR & CR & FR & MU & GTG & – & ZS \\
\citet{Rorseth2024ExplainabilityRALM} & POS & – & PR & TX & PAR & CR & FR & MU & QA & – & ZS \\
\citet{xu-etal-2025-aliice} & EVA & – & PR & TX & SEN & ILC & FR & MU & QA & – & FS \\
\citet{cattan2025doubledipperimprovinglongcontextllms} & APP, EVA & – & IC & TX & PAR & CR & FR & MU & QA & – & \makecell[l]{ZS,\\ FS} \\
\citet{dehghan-etal-2024-ewek} & APP & – & PR & \makecell[l]{TX,\\ GR} & PAR & ILC & FR & MU & QA & – & \makecell[l]{FS,\\ COT} \\
\citet{tahaei-etal-2024-efficient} & APP, RES & – & PR & TX & PAR & ILC & FR & MU & QA & SFT & ZS \\
\citet{xing-etal-2025-evaluating} & EVA, RES & – & \makecell[l]{PR,\\ IC} & TX & PAR & ILC & FR & SI & GTG & – & ZS \\
\citet{aly-etal-2024-learning} & APP & – & PR & TX & PAR & ILC & FR & MU & QA & SFT & – \\
\citet{qi-etal-2024-model} & APP & – & PR & TX & DOC & ILC & FR & MU & QA & – & ZS \\
\citet{byun-etal-2024-reference} & \makecell[l]{APP, EVA,\\ RES} & PLL & – & TX & DOC & \makecell[l]{ILC,\\ CR} & FR & SI & \makecell[l]{GTG,\\ RWG} & – & ZS \\
\citet{zhang-etal-2024-towards-fine-grained} & EVA & – & – & TX & – & ILC & FR & MU & QA & – & – \\
\citet{patel-etal-2024-towards} & APP, RES & – & \makecell[l]{PR,\\ PG} & TX & \makecell[l]{DOC,\\ PAR} & ILC & FR & MU & QA & \makecell[l]{SFT,\\ PT} & – \\
\citet{cao-wang-2024-verifiable} & EVA, RES & – & IC & TX & DOC & ILC & FR & SI & QA & SFT & ZS \\
\citet{buchmann-etal-2024-attribute} & EVA, RES & – & \makecell[l]{PR,\\ PG} & TX & \makecell[l]{PAR,\\ SEN} & CR & FR & MU & \makecell[l]{QA,\\ SUM,\\ FV} & SFT & ZS \\
\citet{xia-etal-2025-ground} & APP, RES & – & PR & TX & PAR & \makecell[l]{CR,\\ NC} & FR & SI & QA & SFT & \makecell[l]{ZS,\\ FS} \\
\citet{Xia2025RALMSelfReasoning} & APP & – & PR & TX & PAR & ILC & IT & MU & QA & SFT & \makecell[l]{ZS,\\ FS} \\
\citet{Dani2024SemioLLM} & \makecell[l]{EVA, RES,\\ APL} & PLL & – & TX & DOC & CR & FR & SI & GTG & – & \makecell[l]{ZS,\\ FS,\\ COT,\\ SC} \\
\citet{laban-etal-2024-summary} & EVA, RES & – & PR & TX & DOC & ILC & FR & MU & SUM & – & ZS \\
\citet{sahinuc-etal-2024-systematic} & EVA, RES & – & IC & TX & PAR & ILC & FR & SI & CTG & – & \makecell[l]{FS,\\ COT,\\ RP} \\
\citet{BittonGuetta2024VisualRiddles} & \makecell[l]{APP, EVA,\\ RES} & – & IC & TX & DOC & ILC & FR & SI & QA & – & ZS \\
\citet{Rajapaksha2025RAGCyberAttack} & RES, APL & PLL & PR & TX & DOC & CR & FR & SI & QA & – & \makecell[l]{ZS,\\ FS} \\
\citet{Shen2024Citekit} & EVA & PLL & \makecell[l]{PR,\\ PG} & TX & PAR & ILC & FR & MU & QA & – & \makecell[l]{ZS,\\ FS,\\ COT,\\ SC} \\
\citet{Sui2024IterativeVerificationAttribution} & APP & – & PR & TX & PAR & ILC & FR & MU & QA & – & FS \\
\citet{hashemi-etal-2024-llm} & EVA & – & – & TX & DOC & – & – & MU & GTG & – & – \\
\citet{huang-etal-2024-learning} & APP & DC & PR & TX & PAR & ILC & FR & MU & QA & \makecell[l]{SFT,\\ RL} & ZS \\
\citet{li-etal-2024-making} & APP, RES & – & IC & TX & DOC & ILC & FR & MU & QA & SFT & \makecell[l]{COT,\\ AO,\\ COC,\\ COQ} \\
\citet{Allen2024QModuleBot} & APL & – & PR & TX & PAR & PAS & FR & SI & QA & – & ZS \\
\citet{nishimura-etal-2024-toward} & \makecell[l]{APP, EVA,\\ RES} & – & IC & TX & DOC & ILC & FR & SI & RWG & – & ZS \\
\citet{Djeddal2024AttributedInformationRetrieval} & EVA & – & \makecell[l]{PR,\\ PG} & TX & DOC & ILC & FR & MU & QA & – & ZS \\
\citet{cohen-wang2024contextcite} & APP & – & IC & TX & \makecell[l]{SEN,\\ TOK} & ILC & FR & SI & \makecell[l]{QA,\\ SUM} & – & ZS \\
\citet{gilson2024enhancinglargelanguagemodels} & APL & PLL & PR & TX & DOC & CR & FR & MU & QA & – & ZS \\
\citet{ramu-etal-2024-enhancing} & APP & – & PG & TX & SEN & ILC & FR & MU & QA & – & FS \\
\citet{penzkofer-baumann-2024-evaluating} & \makecell[l]{APP, EVA,\\ RES} & – & PR & TX & DOC & ILC & FR & MU & QA & \makecell[l]{SFT,\\ RL} & ZS \\
\citet{jung2024evaluatingimpactspecializedllm} & EVA & PLL & IC & TX & DOC & ILC & FR & – & QA & – & – \\
\citet{Zhang2025LitFM} & APP, RES & – & PR & TX & DOC & ILC & FR & MU & \makecell[l]{SUM,\\ RWG,\\ CTG,\\ OTH,\\ OTH,\\ OTH} & SFT & ZS \\
\citet{zhang-etal-2025-longcite} & \makecell[l]{APP, EVA,\\ RES} & – & IC & TX & \makecell[l]{PAR,\\ SEN} & ILC & FR & MU & QA & SFT & \makecell[l]{ZS,\\ FS} \\
\citet{Song2025TrustScore} & APP, EVA & – & PR & TX & DOC & ILC & FR & MU & QA & RL & – \\
\citet{Tilwani2024NeurosymbolicAI} & POS & PLL & IG & TX & DOC & CR & FR & – & GTG & – & – \\
\citet{Hannah2025LLMKG} & APL & PLL & PG & TX & PAR & ILC & FR & SI & QA & – & FS \\
\citet{shaier-etal-2024-adaptive} & EVA, RES & – & IC & TX & DOC & NC & FR & MU & QA & SFT & \makecell[l]{ZS,\\ FS,\\ COT} \\
\citet{huang-etal-2024-advancing} & APP & – & PG & TX & PAR & ILC & FR & MU & QA & \makecell[l]{SFT,\\ SSFT} & ZS \\
\citet{yan2024atomicfactdecompositionhelps} & APP, EVA & – & PG & TX & SEN & CR & FR & MU & QA & SFT & \makecell[l]{FS,\\ COT} \\
\citet{cheng-etal-2025-coral} & EVA, RES & – & PR & TX & PAR & ILC & FR & MU & GTG & SFT & \makecell[l]{ZS,\\ FS} \\
\citet{vladika2024enhancinganswerattributionfaithful} & APP & – & PG & TX & DOC & CR & FR & SI & QA & – & \makecell[l]{ZS,\\ FS} \\
\citet{abolghasemi-etal-2025-evaluation} & EVA & – & PR & TX & PAR & ILC & FR & SI & QA & – & ZS \\
\citet{liu-etal-2024-generation} & APP & – & IC & TX & DOC & ILC & FR & SI & QA & SSFT & – \\
\citet{Qian2025CitationGeneration} & APP, EVA & – & \makecell[l]{PR,\\ PG} & TX & PAR & – & FR & MU & QA & SFT & FS \\
\citet{chang2025scalable} & APP & – & PG & TX & PAR & PAS & FR & SI & OTH & – & – \\
\citet{Venkit2025AEE} & EVA, RES & – & PR & TX & DOC & \makecell[l]{ILC,\\ CR} & FR & MU & QA & – & – \\
\citet{redelaar-etal-2024-attributed} & \makecell[l]{EVA, RES,\\ APL} & – & PR & TX & PAR & CR & FR & MU & QA & – & FS \\
\citet{mathur-etal-2024-matsa} & APP, RES & – & PG & TB & TC & ILC & FR & MU & QA & – & FS \\
\citet{asai2024openscholarsynthesizingscientificliterature} & APP, RES & – & PR & TX & SEN & ILC & FR & SI & QA & SFT & \makecell[l]{ZS,\\ COT} \\
\citet{Gupta2024TREC2024} & APP, RES & – & PR & TX & DOC & ILC & FR & MU & QA & – & \makecell[l]{ZS,\\ FS} \\
\citet{xiao2025quillquotationgenerationenhancement} & EVA, RES & – & PR & TX & SEN & \makecell[l]{NC,\\ QU} & FR & SI & GTG & – & \makecell[l]{ZS,\\ FS,\\ COT} \\
\citet{Hsu2025REC} & APP, RES & – & PR & TX & SEN & \makecell[l]{ILC,\\ CR} & FR & MU & OTH & SFT & ZS \\
\citet{Shetty2024RiskSumm} & APL & – & IC & TX & DOC & ILC & FR & SI & SUM & – & \makecell[l]{ZS,\\ FS,\\ COT} \\
\citet{yang2024darktrustauthoritycitationdriven} & APP & – & IC & TX & DOC & NC & FR & SI & OTH & \makecell[l]{SFT,\\ RL} & ZS \\
\citet{worledge2024extractiveabstractivespectrumuncoveringverifiability} & EVA & PLL & \makecell[l]{PR,\\ PG} & TX & SEN & ILC & FR & MU & QA & – & FS \\
\citet{li-etal-2024-truthreader} & APL & – & PR & TX & DOC & ILC & FR & MU & \makecell[l]{QA,\\ SUM} & SFT & – \\
\citet{soman2024zebrallamacontextawarelargelanguage} & APP & – & PR & TX & DOC & ILC & FR & SI & QA & SFT & ZS \\
\citet{Ateia2024BioRAGent} & APL & – & PR & TX & DOC & ILC & FR & MU & QA & – & FS \\
\citet{zhang-etal-2025-citalaw} & EVA, RES & – & \makecell[l]{PR,\\ PG} & TX & DOC & ILC & FR & MU & QA & SFT & ZS \\
\citet{Wallat2025CorrectnessFaithfulness} & EVA & – & IC & TX & DOC & ILC & FR & MU & QA & – & ZS \\
\citet{Roy2025RAG} & RES, APL & – & PR & TB & TAB & ILC & FR & MU & GTG & – & ZS \\
\citet{Patel2024FactualityOrFiction} & EVA & PLL & – & TX & DOC & NC & FR & MU & QA & – & \makecell[l]{ZS,\\ CA} \\
\citet{Vassos2024RAGVoters} & APL & – & PR & TX & DOC & PAS & FR & SI & QA & – & ZS \\
\citet{sharma-etal-2025-og} & APP & – & PR & \makecell[l]{TX,\\ GR} & PAR & PAS & FR & MU & QA & – & ZS \\
\citet{wu-etal-2025-pa} & APP & – & PR & TX & PAR & ILC & FR & MU & QA & \makecell[l]{SFT,\\ RL} & FS \\
\citet{jin-etal-2025-rag} & EVA, RES & – & PR & TX & SEN & ILC & FR & MU & QA & – & ZS \\
\citet{li-ng-2025-think} & APP & – & IG & TX & PAR & ILC & FR & MU & QA & – & FS \\
\citet{ma-etal-2025-visa} & APP, RES & – & PR & VL & BB & ILC & FR & SI & QA & SFT & ZS \\
\citet{suri-etal-2025-visdom} & APP, RES & – & PR & \makecell[l]{TX,\\ TB,\\ VL} & \makecell[l]{PAR,\\ TAB,\\ IMG} & CR & IT & MU & QA & – & COT \\
\citet{Craig2024Data} & APP & – & PR & TX & PAR & \makecell[l]{ILC,\\ CR} & FR & MU & QA & – & ZS \\
\citet{devine2025aloftragautomaticlocalfine} & APP & – & PR & TX & DOC & – & FR & SI & QA & SSFT & ZS \\
\citet{Ding2025CitationsTrust} & EVA & – & PG & TX & PAR & ILC & FR & MU & QA & – & ZS \\
\citet{he-etal-2025-evaluating-improving} & APP, RES & – & IC & GR & TRI & ILC & FR & SI & OTH & SFT & \makecell[l]{ZS,\\ FS} \\
\citet{Abbas2025Fanar} & APL & – & PG & TX & DOC & – & FR & MU & GTG & \makecell[l]{SFT,\\ RL,\\ PT} & – \\
\citet{chu-etal-2025-towards} & APP & MC & – & \makecell[l]{TX,\\ GR} & SEN & ILC & IT & MU & QA & SFT & \makecell[l]{ZS,\\ FS} \\

\end{longtable}
\setcounter{table}{6}
\begin{center}
\captionof{table}{Overview of the 134 papers on evidence-based text generation with \acp{llm} included in our survey. The annotation follows the multidimensional taxonomy described in Section~\ref{sec:evidence_based_text_generation}. For each paper, we record the Contribution Type, Attribution Approach (Parametric: Par., Non-Parametric: Non-Par.), Citation Characteristics (Citation Modality: Mod., Evidence Level: Evi., Citation Style: Style, Citation Visibility: Vis., Citation Frequency: Freq.), Task, and \ac{llm} Integration (Training: Train., Prompting: Prompt.). Additional information such as evaluation metrics, frameworks, datasets, and benchmarks is available in our public repository (see Appendix~\ref{sec:availability}). Papers are ordered by the publication date of the earliest accessible version, including preprints. For compactness, we employ abbreviations, all of which are defined in Table~\ref{tab:legend}.}
\label{tab:classification_papers}
\end{center}
\end{scriptsize}
\endgroup
\twocolumn

\begin{table*}[t]
\centering
\small
\begin{tabular}{l p{11cm}}
\toprule
\textbf{Evaluation Method} & \textbf{Description} \\
\midrule

Human Evaluation
& Human evaluation metrics use human judges to evaluate different aspects of \ac{llm}-generated texts, typically on numerical scales of 1–5 or 1–10. Most studies define custom evaluation criteria, which limits standardization, comparability, and reusability across approaches. \\
\midrule

Inference-based
& Inference-based metrics use \ac{nli} models to automatically classify whether an \ac{llm}-generated text is entailed by a reference text. This evaluation method is commonly used to determine whether a grounded \ac{llm}-generated text can be attributed to its source and to assess the factual correctness of an \ac{llm}-generated response based on entailment with a ground-truth reference. \\
\midrule

Lexical Overlap
& Lexical overlap metrics evaluate the degree of lexical overlap between an \ac{llm}-generated text and a reference (ground-truth). \\
\midrule

\ac{llm}-as-a-Judge
& \ac{llm}-as-a-judge metrics use \acp{llm} to automatically evaluate the quality of texts generated by other \acp{llm}, thereby eliminating the need for human judges. Typically, the \ac{llm} acting as a judge generates numerical scores, often normalized to a 0–1 scale. \\
\midrule

Retrieval-based
& Retrieval-based metrics quantitatively evaluate how effectively an approach incorporates relevant evidence by comparing the retrieved or integrated evidence with a ground-truth. This method is commonly used to assess whether the correct sources are accurately cited in \ac{llm}-generated texts, as well as to determine whether appropriate sources can be retrieved and attributed to \ac{llm}-generated texts post-hoc. \\
\midrule

Semantic Similarity-based
& Semantic similarity-based metrics assess the degree of semantic similarity between an \ac{llm}-generated text and a reference (ground-truth) text, typically by computing the cosine similarity between their dense vector embeddings. \\
\bottomrule
\end{tabular}

\caption{Explanation of the evaluation methods for evidence-based text generation with \acp{llm}.}
\label{tab:evaluation_method}
\end{table*}

\newcommand{\rowsep}[1]{%
  \noalign{\vskip 0.25ex}%
  \cline{#1}%
  \noalign{\vskip 0.25ex}%
}

\newcommand{\limcell}[1]{%
  \parbox[t]{\wLim}{\setlength{\parskip}{0pt}\setlength{\parindent}{0pt}#1}%
}

\newcommand{\wDim}{0.60cm}
\newcommand{\wMet}{1.3cm}
\newcommand{\wAsp}{3.0cm}
\newcommand{\wTask}{0.7cm}
\newcommand{\wCtx}{2.9cm}
\newcommand{\wLim}{6.4cm}

\newcolumntype{R}[1]{>{\raggedright\arraybackslash}p{#1}}
\newcolumntype{J}[1]{p{#1}}

\newlength{\dashlen}
\settowidth{\dashlen}{\textendash\ }

\newcommand{\dashitem}[1]{%
  \par\hangindent=\dashlen\hangafter=1%
  \noindent\makebox[\dashlen][l]{\textendash\ }#1%
}

\begin{table*}[t]
\centering
\small
\setlength{\tabcolsep}{3pt}
\renewcommand{\arraystretch}{1.08}

\begin{tabular}{@{}
R{\wDim}
R{\wMet}
J{\wAsp}
R{\wTask}
J{\wCtx}
J{\wLim}
@{}}
\toprule
\textbf{Dim.} & \textbf{Metric} & \textbf{Measured Aspect} & \textbf{Tasks} & \textbf{Evaluation Context} & \textbf{Limitations} \\
\midrule

\multirow[c]{1}{*}{ATT}
& Citation NLI
& To what extent do the provided citations support the associated claims?
& QA, GTG, SUM, FV
& sentence-level attribution in long-form text generation
& \limcell{
\dashitem{requires pre-trained NLI model}
\dashitem{assumes each citation sentence represents a single factual claim}
\dashitem{cannot assess uncited claims}
} \\
\rowsep{2-6}

& Auto-AIS-sentence
& To what extent are generated sentences attributable to supporting sources?
& QA, GTG, SUM
& sentence-level attribution in long-form text generation
& \limcell{
\dashitem{requires pre-trained NLI model}
\dashitem{assumes each sentence represents a single factual claim}
\dashitem{requires access to external evidence provided for generation}
\dashitem{does not evaluate citation quality}
} \\
\rowsep{2-6}

& ROUGE
& To what extent does the generated text overlap lexically with the source texts?
& ALL
& summarization quality, lexical overlap-based alignment with sources
& \limcell{
\dashitem{measures lexical similarity rather than attribution}
\dashitem{requires access to external evidence provided for generation}
\dashitem{does not evaluate citation quality}
\dashitem{lacks claim-level granularity for evaluating long-form texts}
} \\
\rowsep{2-6}

& BERT-Score
& To what extent is the generated text semantically similar to the source texts?
& QA, GTG, SUM, CTG, RWG
& summarization quality, semantic similarity-based alignment with sources
& \limcell{
\dashitem{measures semantic similarity rather than attribution}
\dashitem{requires access to external evidence provided for generation}
\dashitem{does not evaluate citation quality}
\dashitem{lacks claim-level granularity for evaluating long-form texts}
} \\
\rowsep{2-6}

& FActScore
& To what extent are generated atomic facts supported by available sources?
& QA, GTG, FV
& claim-level attribution in long-form text generation
& \limcell{
\dashitem{requires pre-trained NLI model}
\dashitem{requires access to external evidence provided for generation}
\dashitem{evaluates factual precision but not factual recall}
\dashitem{typically requires human annotation for atomic fact decomposition}
} \\

\midrule

\multirow[c]{1}{*}{CIT}
& Citation Retrieval
& To what extent do generated citations match the set of supporting sources?
& QA, SUM
& citation-level retrieval alignment with sources
& \limcell{
\dashitem{requires ground-truth citation annotations}
\dashitem{depends on completeness of the oracle citation set}
\dashitem{does not evaluate whether the cited sources support the generated claims}
} \\
\rowsep{2-6}

& Citation Accuracy
& To what extent do generated citations align with the ground-truth citation texts?
& QA
& citation-level text alignment with sources
& \limcell{
\dashitem{requires ground-truth citation text annotations}
\dashitem{evaluates citation text matching rather than source selection or support}
\dashitem{sensitive to citation text wording differences}
} \\

\midrule

\multirow[c]{1}{*}{COR}
& Exact Match
& To what extent do generated answers exactly match the ground-truth answer texts?
& QA
& exact match for short factoid answers
& \limcell{
\dashitem{evaluates exact text matching rather than semantic correctness}
\dashitem{not applicable to long-form text generation}
} \\
\rowsep{2-6}

& Claim Recall
& To what extent does the generated text entail the factual claims in the ground-truth answer?
& QA, SUM
& claim-level correctness in long-form text generation
& \limcell{
\dashitem{requires pre-trained NLI model}
\dashitem{depends on claim decomposition accuracy}
\dashitem{evaluates claim recall rather than precision}
} \\
\rowsep{2-6}

& Accuracy QA
& To what extent is the generated answer correct with respect to the ground-truth answer?
& QA
& answer-level correctness for short answers
& \limcell{
\dashitem{may depend on answer normalization or extraction procedures}
\dashitem{not applicable to long-form text generation}
} \\
\rowsep{2-6}

& Accuracy NLI
& To what extent can the factuality of generated claims be correctly classified?
& QA, SUM, CTG
& claim-level factuality classification for long-form text generation
& \limcell{
\dashitem{requires pre-trained NLI model}
\dashitem{requires human factuality annotations}
\dashitem{depends on claim decomposition accuracy}
\dashitem{may struggle to detect non-factual claims when counter-evidence is limited}
} \\
\rowsep{2-6}

& BLEU-N
& To what extent does the generated text overlap lexically with the ground-truth text?
& QA, GTG, FV
& answer-level lexical overlap-based alignment
& \limcell{
\dashitem{evaluates lexical similarity rather than factual correctness}
\dashitem{less reliable for long-form text generation}
} \\

\bottomrule
\end{tabular}

\caption{Comparative overview of evaluation metrics for core dimensions. The table summarizes frequently reused evaluation metrics identified in Figure~\ref{fig:evaluation}, organized by the three core evaluation dimensions (Dim.), namely attribution, citation, and correctness. For each metric, we describe the measured aspect, applicable task settings, evaluation context, and known limitations. While Figure~\ref{fig:evaluation} provides an overview of usage frequency, this table supports informed metric selection. All abbreviations are listed in Table~\ref{tab:legend}.}
\label{tab:metric-comparison}
\end{table*}

\begin{table*}[t]
\centering
\small
\renewcommand{\arraystretch}{1.08}
\setlength{\tabcolsep}{4pt}

\begin{tabularx}{\textwidth}{@{}l c >{\raggedright\arraybackslash}X >{\raggedright\arraybackslash}X l@{}}
\toprule
\textbf{Framework} 
& \textbf{\makecell[l]{No. of\\ Papers}} 
& \textbf{\makecell[l]{Evaluation\\ Dimension}} 
& \textbf{\makecell[l]{Evaluation\\ Method}} 
& \textbf{Source} \\
\midrule

ALCE & 12 & ATT, COR, LQ & IB, LO, SB & \citet{gao-etal-2023-enabling} \\
\rowsep{1-5}

G-Eval & 2 & ATT, LQ, REL & LJ & \citet{liu-etal-2023-g} \\
\rowsep{1-5}

AEE & 1 & ATT, COR, CIT, REL & LJ, RB & \citet{Venkit2025AEE} \\
\rowsep{1-5}

ALiiCE & 1 & ATT, COR, CIT, LQ & IB, LO, SB & \citet{xu-etal-2025-aliice} \\
\rowsep{1-5}

Attributed Information Retrieval 
& 1 
& ATT, COR 
& IB, LO, RB, SB 
& \citet{Djeddal2024AttributedInformationRetrieval} \\
\rowsep{1-5}

Attribution Bias & 1 & COR, CIT & LO, RB & \citet{abolghasemi-etal-2025-evaluation} \\
\rowsep{1-5}

BEGIN & 1 & ATT & IB, LO, SB & \citet{dziri-etal-2022-evaluating} \\
\rowsep{1-5}

BioGen & 1 & COR, CIT, LQ, REL, RET & HE, RB, SB & \citet{Gupta2024TREC2024} \\
\rowsep{1-5}

CitaLaw & 1 & ATT, LQ & LO, SB & \citet{zhang-etal-2025-citalaw} \\
\rowsep{1-5}

CiteKit & 1 & ATT, COR, CIT, LQ & IB, LO, RB, SB & \citet{Shen2024Citekit} \\
\rowsep{1-5}

\ac{llm}-Rubric & 1 & ATT, CIT & LJ & \citet{hashemi-etal-2024-llm} \\
\rowsep{1-5}

QUILL & 1 & COR, CIT, LQ, REL, RET & LJ, RB & \citet{xiao2025quillquotationgenerationenhancement} \\
\rowsep{1-5}

RAG-RewardBench & 1 & COR, CIT, REL & LJ & \citet{jin-etal-2025-rag} \\
\rowsep{1-5}

RAGAS & 1 & ATT, COR, REL, RET & IB, LJ, RB, SB & \citet{es-etal-2024-ragas} \\
\rowsep{1-5}

RAGE & 1 & CIT & RB & \citet{penzkofer-baumann-2024-evaluating} \\
\rowsep{1-5}

REC & 1 & COR, CIT, LQ, REL & LJ & \citet{Hsu2025REC} \\
\rowsep{1-5}

ScholarQABench & 1 & ATT, COR, LQ, REL & IB, LO, LJ & \citet{asai2024openscholarsynthesizingscientificliterature} \\
\rowsep{1-5}

Trust-Score & 1 & ATT, COR & IB, RB & \citet{Song2025TrustScore} \\
\rowsep{1-5}

VisDoMBench & 1 & COR, RET & LO, RB & \citet{suri-etal-2025-visdom} \\

\bottomrule
\end{tabularx}

\caption{Evaluation frameworks ordered by the number of papers reusing them. For each framework, we annotate the evaluation dimensions covered: attribution (ATT), correctness (COR), citation (CIT), linguistic quality (LQ), preservation (PRE), relevance (REL), and retrieval (RET). We additionally annotate the evaluation methods used by each framework: human evaluation (HE), inference-based (IB), lexical overlap (LO), \ac{llm}-as-a-judge (LJ), retrieval-based (RB), and semantic similarity-based (SB). In addition, all abbreviations are listed in Table~\ref{tab:legend}.}
\label{tab:frameworks}
\end{table*}

\begin{table*}[t]
\centering
\small
\renewcommand{\arraystretch}{1.08}
\setlength{\tabcolsep}{4pt}

\begin{tabularx}{\textwidth}{@{}l c l l X@{}}
\toprule
\textbf{Benchmark} & \textbf{\makecell[l]{No. of\\ Papers}} & \textbf{Task} & \textbf{Domain} & \textbf{Source} \\
\midrule

ALCE & 9 & QA & Social Media, Wikipedia & \citet{gao-etal-2023-enabling} \\
\rowsep{1-5}

RAG-RewardBench & 2 & QA & Multi-Domain & \citet{jin-etal-2025-rag} \\
\rowsep{1-5}

AttributionBench & 1 & GTG, QA & Multi-Domain & \citet{li-etal-2024-attributionbench} \\
\rowsep{1-5}

BEGIN & 1 & GTG & \makecell[l]{News, Social Media,\\ Wikipedia} & \citet{dziri-etal-2022-evaluating} \\
\rowsep{1-5}

CitaLaw & 1 & QA & Legal & \citet{zhang-etal-2025-citalaw} \\
\rowsep{1-5}

LAB & 1 & FV, QA, SUM & Multi-Domain & \citet{buchmann-etal-2024-attribute} \\
\rowsep{1-5}

MultiAttr & 1 & QA & Wikipedia & \citet{patel-etal-2024-towards} \\
\rowsep{1-5}

ScholarQABench & 1 & QA & \makecell[l]{Health, Scientific,\\ Wikipedia} & \citet{asai2024openscholarsynthesizingscientificliterature} \\
\rowsep{1-5}

Trust-Align & 1 & QA & Social Media, Wikipedia & \citet{Song2025TrustScore} \\
\rowsep{1-5}

TabCite & 1 & QA & Finance, Wikipedia & \citet{mathur-etal-2024-matsa} \\
\rowsep{1-5}

VisDoMBench & 1 & QA & \makecell[l]{Scientific, Wikipedia} & \citet{suri-etal-2025-visdom} \\
\bottomrule
\end{tabularx}

\caption{Evaluation benchmarks ordered by the number of papers applying them. For each benchmark, we annotate the tasks it has been applied to, including question answering (QA), grounded text generation (GTG), summarization (SUM), and fact verification (FV), as well as the data domains. In addition, all abbreviations are listed in Table~\ref{tab:legend}.}
\label{tab:benchmarks}
\end{table*}

\begin{table*}[t]
\centering
\small
\renewcommand{\arraystretch}{1.08}
\setlength{\tabcolsep}{4pt}

\begin{tabularx}{\textwidth}{@{}l l c l X@{}}
\toprule
\textbf{Task} & \textbf{Dataset} & \textbf{\makecell[l]{No. of \\ Papers}} & \textbf{Domain} & \textbf{Source} \\
\midrule

\multirow{12}{*}{\makecell[l]{Question \\ Answering}}
& ASQA & 26 & Wikipedia & \citet{stelmakh-etal-2022-asqa} \\
& ELI5 & 22 & Social Media & \citet{fan-etal-2019-eli5} \\
& NaturalQuestions & 19 & Wikipedia & \citet{kwiatkowski-etal-2019-natural} \\
& QAMPARI & 11 & Wikipedia & \citet{amouyal-etal-2023-qampari} \\
& HotpotQA & 10 & Wikipedia & \citet{yang-etal-2018-hotpotqa} \\
& ExpertQA & 9 & Web Search & \citet{malaviya-etal-2024-expertqa} \\
& StrategyQA & 7 & Wikipedia & \citet{geva-etal-2021-aristotle} \\
& HAGRID & 5 & \makecell[l]{Wikipedia, Synthetic} & \citet{kamalloo2023hagridhumanllmcollaborativedataset} \\
& TriviaQA & 5 & \makecell[l]{Web Search, Wikipedia} & \citet{joshi-etal-2017-triviaqa} \\
& 2WikiMultiHopQA & 4 & Wikipedia & \citet{ho-etal-2020-constructing} \\
& GenSearch-Verifiability & 4 & Web Search & \citet{liu-etal-2023-evaluating} \\
& PopQA & 4 & Wikipedia & \citet{mallen-etal-2023-trust} \\
\midrule

\multirow{8}{*}{\makecell[l]{Grounded Text \\ Generation}}
& Wizard of Wikipedia & 4 & Wikipedia & \citet{Dinan2019WizardOfWikipedia} \\
& CMU-DoG & 2 & Wikipedia & \citet{zhou-etal-2018-dataset} \\
& TopicalChat & 2 & \makecell[l]{Wikipedia, News, Social Media} & \citet{Gopalakrishnan2019TopicalChat} \\
& WikiBio GPT-3 & 1 & \makecell[l]{Wikipedia, Synthetic} & \citet{manakul-etal-2023-selfcheckgpt} \\
& SciDuet & 1 & Scientific & \citet{sun-etal-2021-d2s} \\
& \ac{llm}-Rubric & 1 & \makecell[l]{Web Search, Synthetic} & \citet{hashemi-etal-2024-llm} \\
& BioKALMA & 1 & Wikipedia & \citet{li-etal-2024-towards-verifiable} \\
& RefGPT-Fact & 1 & Wikipedia & \citet{yang-etal-2023-refgpt} \\
\midrule

\multirow{4}{*}{Summarization}
& GovReport & 2 & Government & \citet{huang-etal-2021-efficient} \\
& SummEval & 1 & News & \citet{fabbri-etal-2021-summeval} \\
& SummHay & 1 & \makecell[l]{News, Synthetic} & \citet{laban-etal-2024-summary} \\
& CNN DailyMail & 1 & News & \citet{nallapati-etal-2016-abstractive} \\
\midrule

\multirow{4}{*}{\makecell[l]{Related Work \\ Generation}}
& Byun-CR & 1 & Scientific & \citet{byun-etal-2024-reference} \\
& Li-Citation-Graph & 1 & Web Search & \citet{li-ouyang-2025-explaining} \\
& RollingEval & 1 & Scientific & \citet{Agarwal2025LitLLMs} \\
& STRoGeNS & 1 & Scientific & \citet{nishimura-etal-2024-toward} \\
\midrule

\multirow{4}{*}{\makecell[l]{Citation Text \\ Generation}}
& S2ORC & 2 & Scientific & \citet{lo-etal-2020-s2orc} \\
& MCG-S2ORC & 1 & Scientific & \citet{Anand2023CitationTextGeneration} \\
& Gu-CCSG & 1 & Scientific & \citet{gu-hahnloser-2024-controllable} \\
& Sahinuç-ACLRelWork & 1 & Scientific & \citet{sahinuc-etal-2024-systematic} \\
\midrule

\multirow{4}{*}{\makecell[l]{Fact \\ Verification}}
& FEVER & 5 & Wikipedia & \citet{thorne-etal-2018-fever} \\
& Min-Biography & 2 & Wikipedia & \citet{min-etal-2023-factscore} \\
& PubHealth & 1 & Health, News & \citet{kotonya-toni-2020-explainable-automated} \\
& HaluEval & 1 & \makecell[l]{Wikipedia, Synthetic} & \citet{li-etal-2023-halueval} \\
\bottomrule
\end{tabularx}

\caption{Frequent datasets per task in evidence-based text generation with \acp{llm}. For each task in Section~\ref{sec:task}, we display the most frequent datasets. The domain specifies the scope of the data, characterized by its source. This table presents only a sample and does not show the complete list of 231 datasets. The complete dataset table is provided in our repository, as described in Appendix~\ref{sec:availability}. Note, both \ac{llm}-Rubric and STRoGeNS contain multiple datasets, which have been combined in this table for the sake of simplicity.}
\label{tab:datasets}
\end{table*}

\end{document}